\documentclass[10pt,twocolumn,letterpaper]{article}
\usepackage[accsupp]{axessibility}

\usepackage{iccv}
\usepackage{times}
\usepackage{algorithmic}
\usepackage{amsmath}
\usepackage{amssymb}
\usepackage{arydshln}
\usepackage{bbm}
\usepackage{bm}
\usepackage{booktabs}
\usepackage[font=normalsize,labelfont=bf]{caption}
\usepackage{cuted}
\usepackage{graphicx}
\usepackage{pifont}
\usepackage{subcaption}
\usepackage{times}
\usepackage{wrapfig}
\usepackage{xcolor}
\usepackage[numbers,sort]{natbib}
\usepackage[pagebackref=true,breaklinks=true,letterpaper=true,colorlinks,bookmarks=false]{hyperref}
\usepackage{footnote}
\makesavenoteenv{tabular}
\makesavenoteenv{table}
\newif\ifdark
\IfFileExists{/Users/vedaldi/.bash_profile}{
\immediate\write18{%
if defaults read -g AppleInterfaceStyle 2>/dev/null;
then echo \\darktrue > /tmp/displaymode.tex;
else echo \\darkfalse > /tmp/displaymode.tex; fi}
\input{/tmp/displaymode.tex}
\ifdark
\definecolor{pcolor}{HTML}{1E1E1E}
\definecolor{tcolor}{HTML}{C5C5C5}
\else
\definecolor{pcolor}{HTML}{FDF6E3}
\definecolor{tcolor}{HTML}{333333}
\fi
\pagecolor{pcolor}
\color{tcolor}
\hbadness=\maxdimen
\vbadness=\maxdimen
\vfuzz=30pt
\hfuzz=30pt
}{} 
\usepackage{adjustbox}
\makeatletter
\renewcommand{\paragraph}{%
  \@startsection{paragraph}{4}%
  {\z@}{.5ex \@plus 1ex \@minus .2ex}{-1em}%
  {\normalfont\normalsize\bfseries}%
}
\makeatother

\def\iccvPaperID{7813}

\newcommand{\methodname}{STiCA}%
\newcommand{\cmark}{\ding{51}}%
\newcommand{\xmark}{\ding{55}}%

\usepackage{cleveref}
\crefname{section}{Sec.}{Sec.}
\Crefname{section}{Sec.}{Sec.}
\crefname{table}{Tab.}{Tab.}
\Crefname{table}{Tab.}{Tab.}
\crefname{figure}{Fig.}{Fig.}
\Crefname{figure}{Fig.}{Fig.}
\newcommand{\ul}[1]{\underline{{#1}}}

\makeatletter
\def\adl@drawiv#1#2#3{%
        \hskip.5\tabcolsep
        \xleaders#3{#2.5\@tempdimb #1{1}#2.5\@tempdimb}%
                #2\z@ plus1fil minus1fil\relax
        \hskip.5\tabcolsep}
\newcommand{\cdashlinelr}[1]{%
  \noalign{\vskip\aboverulesep
          \global\let\@dashdrawstore\adl@draw
          \global\let\adl@draw\adl@drawiv}
  \cdashline{#1}
  \noalign{\global\let\adl@draw\@dashdrawstore
          \vskip\belowrulesep}}
\makeatother
\makeatletter
\renewcommand{\paragraph}{%
  \@startsection{paragraph}{4}%
  {\z@}{0.5em}{-1em}%
  {\normalfont\normalsize\bfseries}%
}
\makeatother

\newlength\myindent
\title{Space-Time Crop \& Attend:\\ Improving Cross-modal Video Representation Learning.}

\author{Mandela Patrick\thanks{Equal contribution.}, Po-Yao Huang$^*$, Ishan Misra, Florian Metze, Andrea Vedaldi\\
Facebook AI Research\\
{\tt\small mandelapatrick,berniehuang,imisra,fmetze,vedaldi@fb.com}
\and
\and
Yuki M. Asano$^*$, João Henriques \\
Oxford University\\
{\tt\small yuki, joao@robots.ox.ac.uk}
}
\iccvfinalcopy

\begin{document}
\maketitle
\begin{abstract}
The quality of the image representations obtained from self-supervised learning depends strongly on the type of data augmentations used in the learning formulation. Recent papers have ported these methods from still images to videos and found that leveraging both audio and video signals yields strong gains; however, they did not find that spatial augmentations such as cropping, which are very important for still images, work as well for videos. In this paper, we improve these formulations in two ways unique to the spatio-temporal aspect of videos. First, for space, we show that spatial augmentations such as cropping do work well for videos too, but that previous implementations, due to the high processing and memory cost, could not do this at a scale sufficient for it to work well. To address this issue, we first introduce Feature Crop, a method to simulate such augmentations much more efficiently directly in feature space. Second, we show that as opposed to na\"{\i}ve average pooling, the use of transformer-based attention improves performance significantly, and is well suited for processing feature crops. Combining both of our discoveries into a new method, Space-Time Crop \& Attend (STiCA) we achieve state-of-the-art performance across multiple video-representation learning benchmarks. In particular, we achieve new state-of-the-art accuracies of $67.0\%$ on HMDB-51 and $93.1\%$ on UCF-101 when pre-training on Kinetics-400. Code and pretrained models are available\footnote{https://github.com/facebookresearch/GDT}.
\end{abstract}

\section{Introduction}\label{s:intro}

\begin{figure}[t]
\begin{center}
\includegraphics[width=0.99\linewidth]{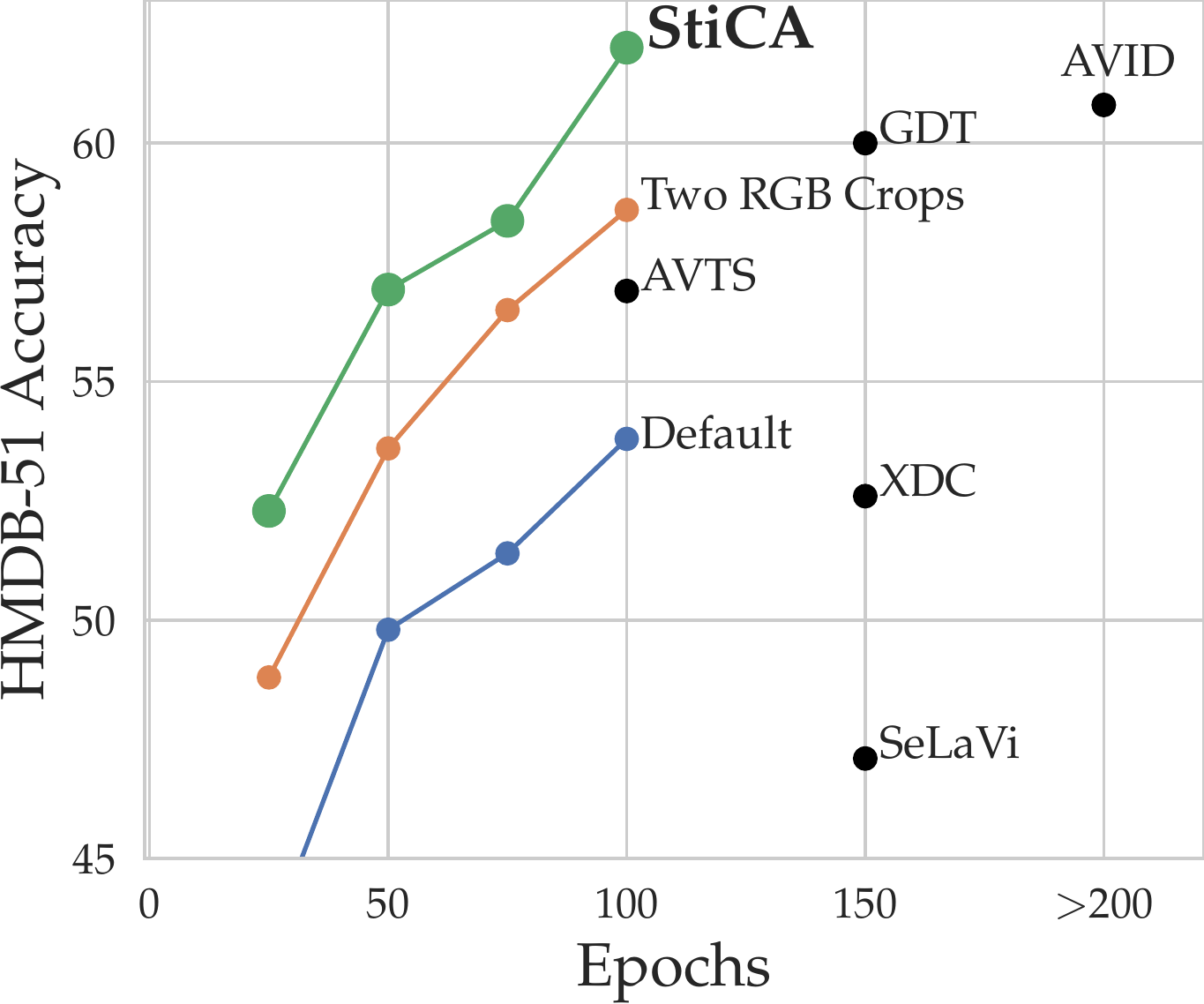}
\vspace{-1em}
\end{center}
   \caption{\textbf{HMDB-51 accuracy vs epoch.} 
   Our method, \textbf{\methodname}, combines space-time crops in feature space with self-attention of time in latent space. 
   This yields significant benefits not only in performance but also in speed compared to cropping in input space using \textit{two RGB crops}, or simply using the \textit{default} cross-modal only loss. 
   Compared to recent state-of-the-art cross-modal self-supervised learning methods 
   (XDC~\cite{alwassel2019self}, GDT~\cite{Patrick2020MultimodalSF}, AVID-CMA~\cite{morgado2020avid}, SeLaVi~\cite{asano2020labelling}) pre-trained on Kinetics-400~\cite{kinetics}
   \methodname{} is able to achieve significantly better results in fewer epochs. 
   \label{fig:splash}
   }
\end{figure}

Visual representations have evolved significantly in the last two decades.
The first generation of representations comprises algorithms such as SIFT~\cite{lowe99object} and HOG~\cite{dalal05histogram} that were designed manually.
The second generation comprises representations learned from data by using deep neural networks and manual supervision~\citep{Krizhevsky12, deng2009imagenet,he16resnet}.
We are now transitioning to the third generation, where representations are learned from data without using any manual annotations by means of self-supervision.
Current self-supervised representations, obtained from methods such as MoCo~\cite{he2019momentum}, SimCLR~\cite{Chen2020ASF} or SwAV~\cite{caron2020swav}, convincingly outperform supervised ones on downstream tasks such as image classification, segmentation and object detection.
Furthermore, most of these methods are based on noise-contrastive instance discrimination, which was proposed in ExemplarCNNs~\cite{dosovitskiy2015discriminative} and put in its current form in~\cite{Wu_2018_CVPR} and~\cite{oord2018representation}.
The idea is to learn representations that are invariant to irrelevant factors of variations, modelled by strong augmentations such as image cropping, while remaining distinctive for the identity of the image.

Noise-contrastive learning is of course not limited to still images.
In particular, a number of recent approaches~\citep{miech2019endtoend, morgado2020avid, Patrick2020MultimodalSF, Han2020SelfsupervisedCF} have used noise-contrastive formulations to learn visual or audio-visual representations.
However, these methods are not as well developed as their counterparts for still images, with current state-of-the-art methods~\citep{Patrick2020MultimodalSF, Han2020SelfsupervisedCF} still lagging behind their supervised counterparts.

In this paper, we identify two areas in which current video representation learning formulations are lacking and improve on them, thus significantly improving upon the current state of the art in this area.

The first shortcoming is the lack of a sufficient encoding of \textbf{spatial invariances}.
For still images, learning spatial invariances has been shown to be one of the most important factors for performance~\citep{caron2020swav, Chen2020ASF}.
Almost all methods achieve some form of spatial invariance simply by applying different spatial augmentations to the images in different epochs of training.
However, learning spatial invariances in this manner requires a slow training process that lasts for many epochs ($\sim$800).
Authors have suggested that packing several augmentations of the same image in a single data batch is more effective as it provides a much stronger and more direct incentive for the network to learn invariances~\cite{caron2020swav}.

For videos, both strategies are less feasible.
Training a model for 200~epochs on Kinetics-400~\cite{kinetics} already requires around $1.5$K GPU hours on recent Nvidia V100 architectures, and with recent datasets such as IG65M~\cite{Ghadiyaram2019} and HowTo100M~\cite{miech19howto100m} only a handful of epochs can realistically be completed.
On the other hand, including multiple augmentations of the same video in a batch rapidly exhausts the memory of GPUs.
Since batch sizes per GPU are already in the single digits due to the size of video data, including several augmentations is unfeasible.
This is particular detrimental for recent contrastive learning approaches such as~\citep{he19momentum,Chen2020ASF}, where reducing the batch size means reducing the pool of negative contrastive samples.

In order to solve this problem, we propose to move spatial augmentations to the feature space, in a manner specifically tailored to contrastive learning.
Instead of extracting a large number $R$ of different augmentations in the input RGB space, we extract only two of them, apply the trunk of the neural network to extract corresponding features, and then extract $R/2$  more augmentations directly in feature space.
In this way, one needs to evaluate the slow and memory taxing feature extraction part of the network only twice, regardless of the number of augmentations that are produced.
We show that this \emph{feature-level augmentation} significantly improves  representation learning performance.

The second challenge that we tackle is how  to best encode \textbf{temporal information} in self-supervised video representation learning.
Currently, most self-supervised video representation learning approaches use 3D-CNNs~\cite{Tran15, Tran18, I3D, xie2017rethinking} that compute convolutions across space and time, but the final representation is generated by na\"{\i}ve global average pooling over space and time, crucially discarding temporal ordering.

In order to address this shortcoming, in this work we propose to use a contextualized pooling function based on the transformer architecture~\cite{vaswani2017attention} for both self-supervised pretraining and supervised finetuning.
The intuition is that, via multi-head self-attention, the transformer can capture temporal dependencies much better than average pooling, especially for longer inputs.
Transformers can also benefit from our feature-level crops, as the latter resemble the common approach of randomly masking the inputs to the transformer~\cite{bert18}.
Experimental results show that this modification improves the performance of the learned video representations substantially, and is cumulative with the benefit of feature crops, at about the same cost of average pooling.

We combine both of our proposed improvements into a new self-supervised learning approach: Space-Time Attention and Cropping (\textbf{\methodname}).
To summarize, with \methodname{} we make the following three main contributions:
\begin{itemize}
\item We demonstrate the benefits of stronger spatial invariances in self-supervised video representation learning for the first time  and we propose feature-level augmentation to implement the latter efficiently.
\item We propose to use transformers to model time more effectively in self-supervised video representations, replacing average as the pooling function.
\item We demonstrate strong performance gains by using the two techniques and obtain state-of-the-art performance on two standard benchmarks ($67.0\%$ on HMDB-51 and $93.1\%$ on UCF-101).
\end{itemize}

\section{Related Works}
\paragraph{Self-supervised Image Representation Learning.}
Self-supervised learning uses pretext tasks to automatically and easily generate differentiable learning signals from the data itself in order to train convolutional neural networks.
A variety of pretext tasks have been proposed such as colorization~\citep{zhang2016colorful, zhang2017split}, predicting artificial rotations~\citep{gidaris2018unsupervised}, in-painting~\citep{pathak2016context}, spatial context~\citep{doersch2015unsupervised, noroozi2016unsupervised}, and clustering features~\citep{caron2018deep, caron2019unsupervised, ji2018invariant, asano2020self, li2020prototypical, caron2020swav}.
Recently, contrastive methods~\citep{Hadsell2006DimensionalityRB, gutmann2010noise} have proven to be particularly effective at learning transferable image representations~\citep{he2019momentum, bachman2019learning, misra2019selfsupervised, Chen2020ASF, grill2020bootstrap, tian2019contrastive, oord2018representation}.

\paragraph{Self-supervised Video Representation Learning.}
For videos, pretext tasks often seek to leverage the temporal dimension to learn representations. 
Such tasks include predicting clip and sequence order~\citep{clip_order, lee2017unsupervised, misra2016shuffle}, future events~\citep{han2019video, Han2020MemoryaugmentedDP}, the arrow of time~\citep{wei2018learning}, 3D geometric transformations~\citep{jing2018self, kim2019self}, playback speed~\citep{benaim2020speednet, Wang2020SelfsupervisedVR, Jenni2020VideoRL, Epstein2020OopsPU}, or motion
statistics~\citep{motion_statistics}. 

\paragraph{Multi-Modal Learning.}
The co-occurrence and synchronicity of multiple modalities from videos have been used to learn visual representations from both audio-video~\citep{Arandjelovic17, owens2018audio, avts, alwassel2019self, morgado2020avid, Patrick2020MultimodalSF, asano2020labelling, ma2020learning}, and speech-video~\citep{miech17learning, sun2019videobert, sun2019contrastive, miech2019endtoend, nagrani2020, li2020learning, korbar2020video, alayrac2020selfsupervised, patrick2020supportset, Stroud2020LearningVR} data. 
Multi-modal representation learning has several practical applications: lip reading~\citep{Chung16a, Chung_2017_lip_reading, afouras2018deep_lip_reading}, audio-visual source separation and localization~\citep{arandjelovic2018objects, zhao2018sound, afouras2018conversation, zhao2019sound, harwath2018jointly, Afouras2020SelfSupervisedLO}, speech recognition~\citep{potamianos2003recent, afouras2018deep}, efficient inference~\citep{gao2019listen, korbar2019scsampler}, egocentric action recognition~\citep{kazakos2019epicfusion} and audio-visual navigation~\citep{chen2019audiovisual}.

\paragraph{Data Augmentations.}
Data augmentation has proven to be useful in training deep learning models in many domains, from vision~\citep{autoaugment, Cubuk2019RandAugmentPD, yun2019cutmix} to speech~\citep{Park2019SpecAugmentAS}.
Data transformations are the foundation of most self-supervised works, and there has been early attempts to even learn the optimal distribution of transformations~\citep{buchler2018improving,autoaugment}. 
Particularly for contrastive learning, the choice of data transformations has been shown to be particularly important to learn desirable invariances and equivariances~\citep{misra2019selfsupervised, Patrick2020MultimodalSF, tian2019contrastive, tian2020makes}.

\paragraph{Transformations in Feature-Space.}
Some works have proposed forms of augmentation in feature-space, by adding noise and linear transformations~\citep{terrance2017dataset}, and by associating samples to prototypes in feature-space~\citep{kuo2020featmatch}. These augmentations do not correspond to interpretable geometric operations, however.
Crops in feature-space are commonly used in supervised detection pipelines, such as Faster R-CNN and region-based architectures~\citep{ren2015faster}, and in earlier detectors based on manually-engineered features~\citep{dalal05histogram}. However, the objective of these transformations is to enumerate a space of outputs (e.g. bounding box predictions) for supervised prediction. 
In self-supervised learning, while~\citep{kalantidis2020hard} uses feature mixing to create harder negatives for contrastive learning, we are instead interested in using feature crop augmentation to achieve spatial invariance.

\paragraph{Temporal Modeling.}
Videos extend images by adding a temporal dimension.
Therefore, there has been a large family of research that has looked into how to model temporal information in videos. 
Early works incorporated temporal information via average pooling of frame/clip-level features~\citep{Sports1M, wang16temporal, Girdhar17}, while later work used 3D convolution neural networks~\citep{Tran15, Tran18, xie2017rethinking} and recurrent-neural networks~\citep{donahue2016longterm}. 
Other approaches leverage long-term temporal convolutions~\citep{varol17}, self-attention~\citep{XiaolongWang18}, relation networks~\citep{abs08496}, multi-scale temporal convolutions~\citep{hussein2019timeception}, or optical flow in a two stream network~\citep{simonyan2014twostream}. 

\paragraph{Transformers in Vision.} 
With the success of the transformer architecture~\citep{vaswani2017attention} in natural language processing~\citep{bert18}, transformers are being used in various vision domains such as image representation learning~\citep{dosovitskiy2020image, chen2020generative, wu2020visual, desai2020virtex, sariyildiz2020learning}, image generation~\citep{parmar2018image}, object detection~\citep{locatello2020objectcentric, carion2020endtoend}, few-shot learning~\citep{doersch2020crosstransformers}, video action recognition~\citep{XiaolongWang18, girdhar2019video,timesformer,neimark2021VTN}, video question-answering~\citep{kant2020spatially}, image-text~\citep{lu2019vilbert, su2019vlbert, Tan_2019, li2019visualbert, tan2020vokenization} and video-text~\citep{sun2019videobert, sun2019contrastive, zhu2020actbert, gabeur2020multimodal, patrick2020supportset, korbar2020video} representation learning. 
\section{Method}\label{s:method}

\begin{figure*}
\centering
\includegraphics[width=0.9\linewidth]{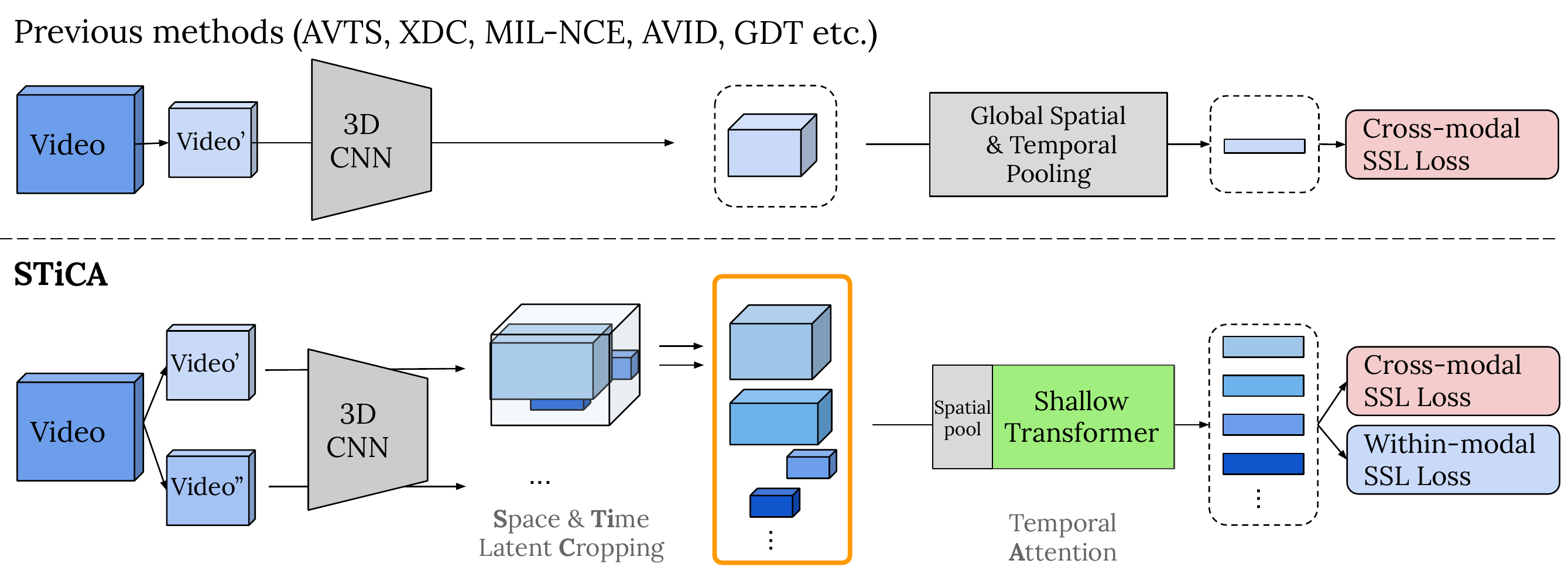}
\vspace{-0.1in}
\caption{\textbf{Approach Overview.} We present a self-supervised approach that learns video representations without labels. \textbf{(Top)} Prior work in video representation learning did not capture spatial invariances, as taking many crops of the input (at varying locations and scales), quickly gets expensive in both compute and memory.
\textbf{(Bottom)} The proposed method generates a large variety of views from only two RGB-crops by cropping in latent space and is particularly tailored to  self-supervised contrastive learning.
The latent crops are essentially masked features, which are 
then further processed by a light-weight temporal transformer. Compared to global pooling, this allows our method to further capture the rich temporal signal.
}
\label{fig:short}
\vspace{-1.0em}
\end{figure*}

\newcommand{\bx}{{\bm x}}
\newcommand{\by}{{\bm y}}
\newcommand{\bz}{{\bm z}}
\newcommand{\bw}{{\bm w}}
\newcommand{\bv}{{\bm v}}
\newcommand{\ba}{{\bm a}}
\newcommand{\bh}{{\bm h}}

Our goal is to learn a general-purpose \emph{data representation} $\Phi : \mathcal{X} \rightarrow \mathcal{Z} = \mathbb{R}^D$ that maps data $\bx\in\mathcal{X}$ to feature vectors $\bz=\Phi(\bx)$.
In the supervised setting, representations are learned end-to-end as components of larger systems that solve certain tasks of interest, such as image or video classification, under the assumption that supervision is available to drive the learning process.
When supervision is not available, representations can still be learned via self-supervision by means of suitable pretext tasks.
Among the latter, \emph{noise contrastive learning} is one of the most popular and successful ones~\citep{oord2018representation,Chen2020ASF}.
We summarize this background next and discuss our extensions in the following sections.

\subsection{Background: Multi-modal contrastive learning}

The idea is to train the representation $\Phi$ to identify data points up to the addition of noise or, more generally, the application of certain nuisance transformations.
To this end, let $g : \mathcal{X} \rightarrow \mathcal{X}$ be transformations sampled in a set $\mathcal{G}$ of possible nuisances (for example random image crops).
Let $\operatorname{sim}(\bz',\bz'')$ be a similarity function comparing representations $\bz'$ and $\bz''$, such as the cosine similarity:
$$
\operatorname{sim}(\bz',\bz'')
=
\frac
{\langle\bz', \bz''\rangle}
{\|\bz'\| \, \|\bz''\|}.
$$
Consider a dataset or batch $\mathcal{B} = \{\bx_1,\dots,\bx_N\}$ of data samples.
Slightly modifying~\citep{Chen2020ASF}, for each sample $\bx_i$, draw a set of random nuisance transformations 
$
\{ g_{\alpha i}\}_{1\leq i \leq N}
$
and let
$
\bz_{\alpha i}= \Phi(g_{\alpha i}(\bx_{i}))
$
be the representations of the transformed samples.
Likewise, consider a second set $\beta$ of transformations
$
\{ g_{\beta i}\}_{1\leq i \leq N}.
$
The noise contrastive loss (NCE) is given by:
\begin{equation}\label{eq:contrastive}
\mathcal{L}(\alpha,\beta)
=
-\frac{1}{N}
\sum_{i=1}^N
\log \frac
{e^{\frac{1}{\tau}\operatorname{sim}(\bm z_{\alpha i}, \bm z_{\beta i})}}
{
\sum_{j=1}^N
e^{\frac{1}{\tau}\operatorname{sim}(\bm z_{\alpha i}, \bm z_{\beta j})}
}
\end{equation}
where $\tau > 0$ is a temperature parameter.
This loss pulls together the representations of samples that only differ by the transformation while pushing apart the others.
Note that this definition is not symmetric in the two arguments $\alpha$ and $\beta$ (i.e., $\mathcal{L}(\alpha,\beta)\not=\mathcal{L}(\beta,\alpha)$.
Note also that we can introduce any number of transformation sets $\alpha,\beta,\gamma,\dots$ and, for each pair, we can obtain a different variant of~\cref{eq:contrastive}.


Recently, works such as~\citep{Patrick2020MultimodalSF} have ported this technique to the video domain
by contrasting modalities.
Each video $\bx=(\bv,\ba)$ consists of a visual component $\bv$ and an audio component $\ba$.
One consider two sets of transformations $g_v$, extracting and augmenting the visual component, and $g_a$, extracting and augmenting the audio component.
We still write $\Phi(g_\alpha(\bx))$ for the feature computed for either visual and audio components, but the symbol means that a modality-specific neural network is applied as needed.\footnote{
In other words, $\Phi = (\Phi_v,\Phi_a)$ is really a pair of networks, producing embedding vectors $\bz_\alpha$ that are compatible regardless of the modality $\alpha \in \{v,a\}$.}

With this, we can derive three variants of~\cref{eq:contrastive}, involving mixed visual-audio and homogeneous visual-visual and audio-audio comparisons.
Their combinations are:
\begin{equation}\label{eq:multi-loss}
\lambda_{va} \mathcal{L}(v, a) +
\lambda_{av} \mathcal{L}(a, v)
+ \lambda_{vv} \mathcal{L}(v_1, v_2) +
\lambda_{aa} \mathcal{L}(a_1, a_2).
\end{equation}
where
$\lambda_{va}$,
$\lambda_{av}$,
$\lambda_{vv}$ and
$\lambda_{aa}$ are non-negative mixing weights.



\paragraph{Challenge 1: Encoding within-modality invariance.}

While all terms in \ref{eq:multi-loss} code for desirable invariances of the representation, several recent papers~\citep{Patrick2020MultimodalSF, morgado2020avid, ma2020learning} have found that the mixed term $\lambda_{va}$ is far more important than the other two; in fact, performance \emph{degrades} if one sets $\lambda_{aa},\lambda_{vv} \neq 0$, meaning that within-modal invariances are not successfully leveraged.
%
Our hypothesis is that within-modality invariance can be beneficial, and that these early negative results are due to the fact that current learning formulations are ineffective at capitalizing on this signal.

As suggested in~\cref{s:intro}, the fact that video data is large means that the batch size used in learning must be small.
As a consequence, a batch can contain only a very small number of different augmentations of the same video sample.
In current multi-modal learning formulations, each video is already transformed twice in order to extract video and audio components, so cross-modal invariance is learned well.
However, the downside is that there is no space left in the batch for multiple visual or audio augmentations.
Thus, within-modality invariance is learned only indirectly --- in particular, as noted in~\cref{s:intro}, two different visual or audio augmentations of the same video are visited by the model only after an entire training epoch.
Next, we address this issue by making it feasible to extract several within-modality transformations in the same batch even for video data.



\subsection{Efficient spatial cropping for augmentation}\label{s:featurecrop}

It has been found that self-supervised learning benefits from, and requires more and stronger augmentations compared to the supervised counterpart for optimal performance~\citep{Chen2020ASF}.
In particular, several papers~\citep{Chen2020ASF,asano2020a,caron2020swav} have suggested that, in the case of still images, the most important type of augmentation is \emph{cropping}. 
Namely, given an RGB image $\bx \in \mathbb{R}^{3  \times H \times W}$ with three channels and height and width $H$ and $W$ respectively, a crop is given by a box $B=(x_\textrm{min},x_\textrm{max},y_\textrm{min},y_\textrm{max})$.
The image tensor is first cropped as:
\begin{equation}\label{e:imcrop}
    C_B(\bx) =
    \bx_{[:,~
    y_\textrm{min}:y_\textrm{max},~
    x_\textrm{min}:x_\textrm{max}]}
\end{equation}
where the $:$ symbol is used to denote an index range.
The cropped tensor is then resized to a tensor
$
\tilde\bx = g(\bx) = R_{H_0W_0}(C_B(\bx)) \in \mathbb{R}^{3\times H_0\times W_0}
$
with a given height and width $H_0\times W_0$.
In practice, $R_{H_0W_0}$ may also apply additional augmentations such as color jittering, as detailed in the experiments.

As for the visual part $\bv \in \mathbb{R}^{3\times T\times H\times W}$ of a video, the situation is similar, except that the video also contains an additional temporal dimension $T$.
To avoid extreme spatial jittering and keep objects aligned, a spatial crop is usually taken at the same location in the input space throughout the whole temporal dimension, so we consider the tube
$
B =
(
    x_\textrm{min},x_\textrm{max},
    y_\textrm{min},y_\textrm{max},
    t_\textrm{min},t_\textrm{max}
)
$
and define
$
\tilde \bv = g_v(\bv) = R_{H_0W_0}(C_B(\bv)) \in \mathbb{R}^{3 \times T_0 \times H_0 \times W_0}
$
by extending~\eqref{e:imcrop} in the obvious way.

The deep neural network $\bz = \Phi(\tilde\bv)$ mapping $\bv$ to its corresponding code $\bz$ is fed with tensors with two spatial dimensions and a temporal one.
Such networks, often called 3D for this reason, include R3D~\citep{hara2017learning}, S3D~\citep{xie2017rethinking} and R(2+1)D~\citep{Tran18}.
As customary in deep convolutional neural networks, they first produce an intermediate tensor with lower space-time resolution and then pool the latter to obtain a single code vector for the entire video.
We explicitly break this down into three functions
\begin{equation}\label{e:decomp}
    \Phi(\tilde\bv) = (\mathcal{P}_t \circ \mathcal{P}_{s} \circ \Psi)(\tilde\bv)
\end{equation}
Here, the first function is a 3D convolutional neural network
$
\Psi(\tilde\bv) \in \mathbb{R}^{D \times T_1 \times H_1 \times W_1}
$
producing a tensor with reduced resolution $T_1<T_0$, $H_1<H_0$, $W_1<W_0$.
The operators $\mathcal{P}_{s}$ and $\mathcal{P}_t$ collapse, respectively, \underline{s}patial and \underline{t}ime dimensions via average pooling.

Now consider implementing term $\mathcal{L}(v_1,v_2)$ in~\ref{eq:multi-loss}.
In this case, one samples from each video $\bx_i$ two different space-time crops $g_{v_1i}(\bx_i)$ and $g_{v_2i}(\bx_i)$, each corresponding to random tubes $B_1$ and $B_2$ respectively.
The tubes are not sampled entirely independently, however, as they have the same temporal extent $(t_\textrm{min},t_\textrm{max})$.

\paragraph{Na\"{\i}ve multiple spatial cropping} 

In practice, \citep{caron2020swav, Chen2020ASF, li2020centerwise, misra2019selfsupervised} show that taking multiple image crops improves self-supervised image representations.
We can achieve a similar effect for videos by summing losses 
$
\mathcal{L}(v_\alpha,v_\beta)
$
for sets of visual transformations $v_\alpha \not= v_\beta$, obtained by sampling multiple space-time tubes for each video, but this is practically difficult, both due to the large memory footprint and the compute overhead of the slow 3D CNN for each crop.

The Multi-Crop approach introduced by SwAV~\citep{caron2020swav} in the image domain combined with our asymmetric contrastive formulation~\eqref{eq:contrastive} can partially reduce the complexity.
For Multi-Crop, we consider three crop sizes $\alpha \in \{L_1,L_2,S\}$ where $L_1$ and $L_2$ stands for large and $S$ for small.
The use of a small crop allows to reduce the memory consumption when the representation $\Phi$ is computed.
We then have losses:
$$
\mathcal{L}(v_{L_1},v_{L_2})
+\mathcal{L}(v_{L_2},v_{L_1})
+\mathcal{L}(v_{L_1},v_{S})
+\mathcal{L}(v_{L_2},v_{S}).
$$
While operating on small videos saves some computation, in practice this approach is insufficient to allow using more than a handful of crops in total.

\paragraph{Efficient cropping in feature space.}

As illustrated in~\cref{fig:short}, a much more efficient alternative to cropping the input video is to crop intermediate features. 

To do so, we first apply the trunk $\Psi$ of the representation to an input-space crop of the visual component of the video $\tilde \bv = R_{H_0W_0}(C_B(\bv))\in\mathbb{R}^{D\times T_1\times H_1\times W_1}$.
Then we can efficiently construct a new view of this data by applying the \textit{Feature Crop} $C_{\bar B}$ directly on each intermediate representation, yielding
\begin{equation}\label{e:featurecrop}
    \bar \bv =
    C_{\bar B}(\Psi(\tilde \bv))  =   \Psi(\tilde \bv)_{[
    t_\textrm{min}:t_\textrm{max},~
    y_\textrm{min}:y_\textrm{max},~
    x_\textrm{min}:x_\textrm{max}]}
\end{equation}
Since the operator $C_{\bar B}$ is lightweight, it can be used to compute several such random views efficiently;
by comparison, cropping the input RGB images requires recomputing the trunk $\Psi$  multiple times.

In practice, given an input video $\bv$, we generate the following views.
First, we apply two crops in RGB space, producing two large crops $L_1$ and $L_2$.
Then, for each of those, we use the operator~\eqref{e:featurecrop} to generate $m$ medium-sized and $n$ small-sized crops
$
\mathcal{T}_i =
\{M_1 L_i, \dots, M_m L_i, S_1 L_i, \dots, S_n L_i\}
$.
We define an overall within-modality loss by summing losses for each pairs of views in $\mathcal{T}$ with exception of pairs where both crops are small:
\begin{multline}\label{e:lossvisual}
L_{vv}
=
\sum_{\alpha,\beta}
\mathcal{L}(v_\alpha, v_\beta)
+
\mathcal{L}(v_\beta, v_\alpha),
 ~~~ \text{where}
\\
(\alpha,\beta) \in 
(\mathcal{T}_1\times \mathcal{T}_2) - (\mathcal{S}_1\times \mathcal{S}_2)
\end{multline}
Note that there are $2((m+n)^2-n^2)$ terms in this loss.
This is a far greater number of comparison than afforded by the two initial input-space RGB crops.

\subsection{Temporal modelling with transformers}\label{s:meth:transformer}


We now discuss our second improvement: better modelling of time.

\paragraph{Challenge 2: Modelling time better.} 

Contrary to spatial invariance, models should not be fully invariant to time as the latter can encode causality and with it semantics: a video of someone starting a fire is very different from its reversed version, in which someone extinguishes it.
In standard 3D networks, features in the trunk are sensitive to the temporal order, but this information is lost in the final stage, where temporal averaging is applied.
We argue that the value of the lost information increases with the length of the video, and that this information can be leveraged by switching to a different pooling function.

\paragraph{Temporal transformer.}

We propose to tackle this issue by replacing average pooling in time $\mathcal{P}_t$ in~\cref{e:decomp} with a transformer $\mathcal{P}_\text{transf}$.
Transformers~\citep{vaswani2017attention} have been shown effective for representing sequential inputs in the NLP domain~\citep{bert18, raffel2019exploring, lewis2019bart, radford2019language}.
After spatial averaging, the output $\bh=\mathcal{P}_s(\Psi(\tilde \bv)) \in \mathbb{R}^{D\times T_1}$ of the network has one feature vector per time step, and is thus amenable to processing by a transformer.
The feature $\bh$, which differs in latent time-dimension size from its uncropped variant can be seen as masking the transformer's attention.
Masking attention has been used in transformer encoder-decoder training to prevent the model from cheating~\citep{devlin18bert:} and encourage it to leverage information from the context.
We use a shallow and light-weight transformer on top of our feature cropping procedure, which we show to be sufficient to reap the benefit of better temporal modelling incurring only a very small computational cost.
We use 2-layers and 4 self-attention heads and provide further details on the transformer architecture in the Appendix.


\subsection{Overall loss}
Our combined model, \textbf{\methodname}, better learns space-time invariances and relationships by cropping in space-time and leveraging  temporal attention with a transformer.
For training, we sample $N$ videos in a batch and, for each of them, compute two `large' visual crops in RGB space, $2(n+m)$ small and medium feature crops (\cref{s:featurecrop}), and an audio augmentation $a$.
With those, the overall objective is obtained by summing the within-modality loss $L_{vv}$ from~\cref{e:lossvisual} to the cross modality losses:
\begin{align}
L
& =
\lambda_{vv} L_{vv} + \lambda_{va} L_{va}, \label{loss:total}
\end{align}
where $L_{va} {=} 
\mathcal{L}(v_{L_1}, a)
+
\mathcal{L}(v_{L_2}, a)
+
\mathcal{L}(a,v_{L_1})
+
\mathcal{L}(a,v_{L_2})
$.

\section{Experiments}\label{s:experiments}

\begin{table*}[tb]
\footnotesize
\caption{\textbf{Comparison experiments and ablations.} We compare key parameters and settings of our proposed method. We report results model performance at epoch 100 and with 30 frames and without transformer unless noted otherwise.
}\label{tab:ablation}
\setlength{\tabcolsep}{3pt}
	\begin{subtable}[t]{.58\linewidth}\centering
		{
\footnotesize
\begin{tabular}{llcc}
\toprule
Cropping-strategy                    & Resolution                               & GPU-h/epoch & Acc. \\
\midrule
Default                              & $1\!\times \!112^2$                      & $17.3$      & $54.0$ \\
Two RGB Crops                        & $2\!\times \!112^2$                      & $29.3$      & $58.6$ \\
Multi RGB Crops~\cite{caron2020swav} & $2\!\times \!112^2$ + $1\!\times \!96^2$ & $46.7$      & $59.3$ \\
Ours (Feature Crop)                  & $2\!\times \!112^2$  + latent            & $29.3$      & $\bf{60.4}$ \\
\bottomrule
\end{tabular}
\vspace{0.5em}
\caption{\textbf{Cropping} yields benefits but requires more compute.
Our feature crops are efficient \textit{and} outperform~\cite{caron2020swav}.
Note that all models are trained for 100 epochs.%
\label{tab:ablation:spatial_crop}}
\vspace{-1em}}	
	\end{subtable}
	\hfill
	\begin{subtable}[t]{.39\linewidth}\centering
		{
\begin{tabular}{c cc cc}
\toprule
\multicolumn{2}{c}{\underline{$l$-Spatial size}} & \multicolumn{2}{c}{\underline{$l$-Temporal size}} & Acc. \\
$M$ & $S$ & $M$ & $S$ \\
\midrule
$1\!\times7^2$        &                       & $1\!\times4$        &                   & $54.0$\\
$1\!\times \!6^2$  & $\! 2\!\times \!4^2$  & $1\!\times4$         &                  & $59.9$ \\ 
$1\!\times \!6^2$  & $\!2\!\times \!4^2$   & $2\!\times \!3$        & $\! 1\!\times \!2$ & $58.4$ \\
$2\!\times \!6^2$  & $\!4\!\times \!4^2$   & $2\!\times \!3$        & $\! 1\!\times \!2$ & $\bf{60.4}$ \\
\bottomrule
\end{tabular}
\caption{\textbf{Feature crops.} Heavier augmentations in latent~($l$) space and time lead to better representations.\label{tab:ablation:feature_crop}}
}	
	\end{subtable}
	\\
	\begin{subtable}[t]{.29\linewidth}\centering
	    {\begin{tabular}{llc}
\toprule
Pretraining  & Finetuning & Acc \\
\midrule
$\mathcal{P}_t$             & $\mathcal{P}_t$             & $54.0$ \\
$\mathcal{P}_t$             & $\mathcal{P}_\text{transf}$ & $54.6$ \\
$\mathcal{P}_\text{transf}$ & $\mathcal{P}_t$             & $52.1$ \\
$\mathcal{P}_\text{transf}$ & $\mathcal{P}_\text{transf}$ & $\bf{60.3}$ \\
\bottomrule
\end{tabular}
\caption{\textbf{Pooling.} Compared to Average-Pooling ($\mathcal{P}_t$), Transformer-based pooling ($\mathcal{P}_\text{transf}$) gives stronger performance.\label{tab:ablation:pooling}}}	
	\end{subtable}
	\hfill
	\begin{subtable}[t]{.33\linewidth}\centering
		{
\begin{center}
\begin{tabular}{ccccc}
\toprule
Transf.? & Layers & Params & GFLOPS & Acc. \\
\midrule
\xmark & $0$ & $37.2$M & $77.7$ & $54.0$ \\
\xmark & $0$ & $42.8$M & $80.0$ & $57.3$ \\
\cmark & $2$ & $42.4$M & $77.8$ & $\bf{60.3}$ \\
\cmark & $4$ & $47.7$M & $77.8$ & $58.3$ \\
\bottomrule
\end{tabular}
\end{center}
\vspace{-1em}
\caption{\textbf{Architecture.} Using up to two transformer layers gives gains, not due to more trainable parameters.\label{tab:ablation:num_layers_heads}}
}	
	\end{subtable}
	\hfill
	\begin{subtable}[t]{.31\linewidth}\centering
		{
\begin{tabular}{cccc}
\toprule
$C_\mathrm{space}$ &  $C_\mathrm{time}$  & T?  & Acc. \\
\midrule
{\color{gray}\xmark} & {\color{gray}\xmark} & {\color{gray}\xmark} & $54.0$ \\
\cmark  & {\color{gray}\xmark} & {\color{gray}\xmark} & $59.9$ \\
\cmark  & \cmark & {\color{gray}\xmark} & $60.4$ \\
\cmark  & \cmark & \cmark & $\bf{62.0}$\\
\bottomrule
\end{tabular} 
\caption{\textbf{Combined gains.} Feature crop in space $C_\mathrm{space}$ and time $C_\mathrm{time}$ and transformer pooling (T) add cumulative benefits. \label{tab:ablation:cumul_gains}}
}	
	\end{subtable} \\ 
	\begin{subtable}[t]{.98\linewidth}\centering
		{
    \begin{tabular}{c ccc | ccc | cc}
        \toprule
        Method &  \multicolumn{3}{c}{\textbf{RGB-Crops}} & \multicolumn{3}{c}{\textbf{Multi-scale RGB-Crops}} & \multicolumn{2}{c}{\textbf{Feature Crops}}\\
        & 1x & 2x & 4x{\color{red}$^*$}  & 2x112 + 1x96 & 2x112 + 2x96 & 2x112 + 6x96{\color{red}$^*$} & (1x7, 1x4) & (2x6 + 4x4, 2x3 + 1x2) \\
        \midrule
        GPU-h/epoch & 17.3 & 29.3 & 60.0 & 46.7 & 53.3 & 100.7 & 29.3 & 30.0  \\
        \bottomrule
    \end{tabular} 
\caption{\textbf{Speed.} Input-crops are slow: {\color{red}$*$} methods require reducing batch sizes (see Appendix) as activations do not fit on GPU. \label{tab:ablation:cropvstime}}
}	
	\end{subtable}
\vspace{-2.0em}
\end{table*}

We first describe the datasets (\cref{s:datasets}) and implementation details (\cref{s:impl_details}) for pretraining.
In \cref{s:downstream}, we describe the downstream tasks for evaluating the representation obtained from self-supervised learning.
In~\cref{s:ablation}, we ablate the various components of our method, and the importance of temporal context and multi-modality in \cref{s:findings}.
Lastly, in \cref{s:sota}, we compare with prior work in video and multi-modal representation learning.

\paragraph{Data.\label{s:datasets}}

We pretrain on the Kinetics-400 dataset~\citep{kinetics}, which contains about 230K training videos and 13K validation videos belonging to 400 action classes.
This dataset is the ``ImageNet'' for video representation learning due to its moderate size and being public, allowing for broad access and comparability.
After pretraining, we evaluate using video action retrieval and action recognition on HMDB-51~\citep{HMDB51} and UCF-101~\citep{UCF101}.
HMDB-51~\citep{HMDB51} consists of 7K video clips spanning 51 different human activities.
HMDB-51 has three train/test splits of size 5K/2K respectively.
UCF-101~\citep{UCF101} contains 13K videos from 101 human action classes, and has three train/test splits of size 11K/2K respectively.

\paragraph{Implementation details.\label{s:impl_details}}

Following~\citep{Patrick2020MultimodalSF}, we use the R(2+1)-18~\citep{Tran18} network as visual encoder and ResNet~\citep{he16resnet} with 9 layers as audio encoder. 
We train for $100$ epochs and use $30$ frames with temporal stride of 1 at sampling rate of 30fps at spatial resolution of $112\times112$ as input.
In our ablations, we evaluate the learned representation by finetuning the visual encoder on fold 1 of the HMDB-51~\citep{HMDB51} action recognition dataset.
Further implementation details are given in the Appendix.

\subsection{Downstream tasks\label{s:downstream}}


\paragraph{Video action retrieval.}

For video retrieval, we follow the standard protocol described in~\citep{clip_order}.
We use the split 1 of UCF-101, and additionally HMDB-51.
We uniformly sample $10$ clips per video, max pool and then average the features after the last residual block for each clip per video.
We use these averaged features from the validation set to query the videos in the training set.
If the class of a retrieved video matches the class of query video, we count it as a match.
We measure recall at $k{=}1,5,20$.

\paragraph{Video action recognition.}

As is standard in the literature, we evaluate our pretrained representations by finetuning our visual backbone on the video action recognition task on HMDB-51 and UCF-101 datasets.
We closely follow the finetuning schedule of GDT~\citep{Patrick2020MultimodalSF}.
During finetuning, we use SGD with initial learning rate $0.0025$, which we gradually warm up to $0.02$ in the first $2$ epochs
The weight decay is set to $0.005$ and momentum to $0.9$.
We use a mini-batch size of $32$ and train for $12$ epochs with the learning rate multiplied by $0.05$ at $6$ and $10$ epochs.
For training, we randomly sample $1$s clips per video, and during evaluation, we uniformly sample $10$ clips from each video and apply 3-crop evaluation as in~\citep{slowfast}.

\subsection{Comparison experiments and ablations\label{s:ablation}}

\paragraph{Cropping augmentation.}

In~\Cref{tab:ablation:spatial_crop}, we ablate the importance of spatial augmentation in learning video representations.
We compare our proposed Feature Crop augmentation, $C_{\bar B}$, to the recently proposed Multi-Crop augmentation strategy~\citep{caron2020swav} and other baseline approaches.
Multi-Crop has proven to be effective in image self-supervised learning because it forces the model to learn local-to-global associations, by explicitly enforcing invariance between features of large-crops and those of multiple small crops.
While effective, it can be particularly computationally intensive, which, with our hardware, limits its use to only two large crops and one small crop when applied to video representation learning.
Our proposed Feature Crop is not only more efficient, but outperforms Multi-Crop by $1.1\%$ when the learned representations is applied to action classification in HMDB-51.
By cropping in feature space, we achieve a similar effect but can increase the number of small crops from 1 to 6 without increasing compute time.

\paragraph{Feature crop parameters.}

In~\Cref{tab:ablation:feature_crop}, we study the parameters of our Feature Cropping approach.
We find that even our basic variant, which does one medium $6 \! \times \!6$ crop and two $4\! \times \! 4$ small crops (by cropping a $7\! \times \!7$ tensor) increases performance by nearly $6\%$, which is a relative improvement of more than $10\%$.
If we further increase the number of crops in time and space, the performance increases from $59.9\%$ to $60.4\%$.

\paragraph{Pooling Function.}

In~\Cref{tab:ablation:pooling}, we test temporal aggregation.
We find that using a shallow transformer significantly outperforms simple average pooling by more than $5\%$; however, transformer pooling must be used both for pre-training the representation and for finetuning it on the target dataset.

\paragraph{Transformer architecture.}

In~\Cref{tab:ablation:num_layers_heads}, we test variants of the transformer architecture, including ablating iit altogether.
We find that temporal modelling as measured by downstream performance peaks at two layers, likely due to optimization difficulties of deeper transformers with SGD\@.
We also compare to a model with approximately the same number of parameters as our 2-layer transformer (achieved by increasing the networks' last block's hidden dimension to $640$).
We find that the transformer still yields gains of $3\%$, indicating that it not the number of parameters but the modelling of time that is crucial for strong performance.

\paragraph{Combining Feature Crops and Transformer Pooling.}

In~\Cref{tab:ablation:cumul_gains}, we show that combining Feature Crops in space and time, and then adding transformer pooling yield additive gains, with the best result obtained by combining all effects (which corresponds to \textbf{\methodname}).
This shows that space-time augmentations and transformer pooling are complementary.

\paragraph{Cropping efficiency.}

In~\Cref{tab:ablation:cropvstime}, we compare training times (normalized to GPUs$\times$hours) for Kinetics-400 epochs for the various spatial crops considered.
We make two observations: First, the compute cost of RGB crops scales proportionally to their number because a full forward pass is required for each crop. Second, using a larger number of RGB crops eventually requires to decrease the batch size, which increases significantly the training time.
In contrast, the cost of Feature Crop remains roughly constant no matter the number of crops.

\begin{table}
\footnotesize
\centering
\makebox[0pt][c]{\parbox{0.48\textwidth}{%
\begin{minipage}[b]{0.48\hsize}\centering
\setlength{\tabcolsep}{3pt}
\begin{tabular}{c c c c c}
\toprule
\multicolumn{2}{c}{\underline{{Frames}}} &   & \multicolumn{2}{c}{\underline{{Accuracy}}} \\
Pretrain & Finetune & & GAP    & Transf.       \\
\midrule
30       & 30       & & $54.0$ & $\bf{60.3}$  \\
60       & 60       & & $62.4$ & $\bf{66.1}$ \\
90       & 90       & & $58.0$ & $\bf{66.9}$\\
\bottomrule
\end{tabular}
\vspace{-0.5em}
\caption{\textbf{Temporal context.} We report results with different number of frames on finetuning accuracy.\label{tab:ablation:transformer}}

    \vspace{-1em}
\end{minipage}
\hfill
\begin{minipage}[b]{0.42\hsize}\centering
%
\setlength{\tabcolsep}{3pt}
\begin{tabular}{cccc}
\toprule
$\lambda_{va}$ &  $\lambda_{vv}$  & F. Crop?  & Acc. \\
\midrule
0    & 1   & No  & $43.3$ \\
1    & 0   & No  & $54.0$ \\
0.5  & 0.5 & No  & $58.6$ \\
0.5  & 0.5 & Yes & $\bf{60.3}$\\
\bottomrule
\end{tabular} 
\vspace{-0.5em}
\caption{\textbf{Loss.} Combining within-modal and cross-modal loss with Feature-crops is key. \label{tab:ablation:loss}}

    \vspace{-1em}
\end{minipage}
}}
\end{table}

\subsection{Temporal Context and Multi-modality\label{s:findings}}

\paragraph{Length of temporal context.}

In~\Cref{tab:ablation:transformer}, we show the importance of leveraging longer context to improve video self-supervised representation learning.
Similar to the supervised regime~\citep{varol17, XiaolongWang18}, we observe improved accuracy as we increase the number of frames used during pretraining and fine-tuning.
More importantly, the transformer pooling layer is better able to exploit this additional context, outperforming average pooling by over $4\%$ for all frame lengths.
Notably, there is a drop in performance when using GAP for extremely long contexts (90 frames).

\setlength{\tabcolsep}{4pt}
\begin{table}[t]
\footnotesize
\begin{center}
\begin{tabular}{l l l  c c }
\toprule
	\textbf{Method}                              & \textbf{Architecture} & \textbf{Dataset} &
	\multicolumn{2}{c}{\textbf{Top-1 Acc\%}} \\
				                               &  &      & HMDB & UCF       \\
\midrule
	Supervised                                & R(2+1)D-18  & K-400 & 70.4  & 95.0 \\
\midrule
	SeLaVi~\cite{asano2020labelling}          & R(2+1)D-18  & K-400 & 47.1 & 83.1  \\
	TempTrans~\cite{Jenni2020VideoRL}     & R3D-18      & K-400 & 49.8 & 79.3  \\
	PEMT~\cite{lee2021parameter} & SlowFast & K-400 & - & 85.2 \\
	XDC~\cite{alwassel2019self}               & R(2+1)D-18  & K-400 & 52.6 & 86.2  \\
	MemDPC~\cite{Han2020MemoryaugmentedDP}    & R-2D3D      & K-400 & 54.5 & 86.1 \\
	AVSF~\cite{xiao2020audiovisual}    & AVSF  & K-400 & 54.6 & 87.0  \\
	AVTS~\cite{avts}                          & MC3-18      & K-400 & 56.9 & 85.8  \\
	CPD~\cite{li2020learning}                 & R3D-50      & K-400 & 57.7 & 88.7  \\
	AVID~\cite{morgado2020avid}               & R(2+1)D-18  & K-400 & 60.8 & 87.5  \\
	GDT~\cite{Patrick2020MultimodalSF}        & R(2+1)D-18  & K-400 & 60.0 & 89.3  \\
	ACC~\cite{ma2020learning}                 & R3D-18      & K-400 & 61.8 & 90.2  \\
	GLCM~\cite{ma2021contrastive}              & R3D-18      & K-400 & 61.9 & 91.2  \\
	CoCLR~\cite{Han2020SelfsupervisedCF}      & S3D         & K-400 & 62.9 & 90.6  \\
	CVLR~\cite{Qian2020SpatiotemporalCV}\footnote{Concurrent work.}      & R3D-50      & K-400 & 66.7 & 92.2 \\ 
\midrule
	\bf{Ours: \methodname}                                 & R(2+1)D-18  & K-400 & \bf{67.0} &\bf{93.1}  \\
\midrule
\midrule
\color{gray}
    SeLaVi~\cite{asano2020labelling}         & R(2+1)D-18 & VGGS    & 53.1      & 87.7\\
    Speech2Act~\cite{nagrani2020}            & S3D-G      & Movie   & 58.1      & -- \\
    DynamoNet~\cite{Diba_2019}               & ResNext101 & Y8M     & 58.6      & 87.3\\
	MIL-NCE~\cite{miech2019endtoend}         & S3D        & HT      & 61.0      & 91.3\\
    AVTS~\cite{avts}                         & MC3-18     & AS      & 61.6      & 89.0\\
	AVID~\cite{morgado2020avid}              & R(2+1)D-18 & AS      & 64.7      & 91.5\\
	Textual~\cite{Stroud2020LearningVR}      & S3D-G      & WVT-70M & 65.3      & 90.3 \\
	GDT~\cite{Patrick2020MultimodalSF}       & R(2+1)D-18 & AS      & 66.1      & 92.5\\
	\cdashlinelr{1-1}
	ACC~\cite{ma2020learning}                & R(2+1)D-18 & AS      & 67.2      & 93.5  \\
	ELo~\cite{piergiovanni2020evolving}      & R(2+1)D-50 & Y2M     & 67.4      & 93.8\\
	XDC~\cite{alwassel2019self}              & R(2+1)D-18 & IG65M   & 68.9      & \bf{95.5}\\
	GDT~\cite{Patrick2020MultimodalSF}       & R(2+1)D-18 & IG65M   & {72.8}      & 95.2\\
	MMV~\cite{alayrac2020selfsupervised}     & TSM-50x2   & AS+HT   & \bf{75.0} & 95.2\\
\bottomrule
\end{tabular}
\end{center}
\vspace{-1.5em}
\caption{\textbf{Comparison to SoTA for action recognition}. Dashed line indicates position of our Kinetics-400 model in comparison to models trained with many more videos. We follow standard evaluation protocol across 3-folds. For linear evaluation results see~\cref{tab:sota_linear}. \label{tab:sota_full_transfer}}
\end{table}

\paragraph{Loss.}

Lastly, in~\Cref{tab:ablation:loss}, we study the effect of combining multi-modal learning signals with our contributions.
In the first row, we have the baseline of naively extending SimCLR~\citep{Chen2020ASF} to the video domain, by learning invariances to spatial augmentations of two large-crops.
Compared to this, the cross-modal baseline (row 2) already achieves gains of more than $10\%$.
While adding a within-modal invariance adds another 4.6\%, we find that the best performance is obtained with our feature crops, adding another 1.7\% in performance and showing its unique potential to supplement cross-modal signals.

\subsection{Comparison with the state of the art\label{s:sota}}

\paragraph{Video Action Recognition.}

In~\Cref{tab:sota_full_transfer}, we evaluate our pretraining approach on the standard HMDB-51 and UCF-101 action recognition benchmarks after pretraining on the Kinetics-400 dataset. 
Firstly, we find our model outperforming the similar NCE-based GDT~\citep{Patrick2020MultimodalSF} model by $7.0\%$ and $3.8\%$ on HMDB-51 and UCF-101.
We further significantly outperform the current state-of-the art methods \mbox{CoCLR}~\citep{Han2020SelfsupervisedCF} by $4.1\%$ and $2.5\%$ and CVLR~\citep{Qian2020SpatiotemporalCV} by $2.6\%$ and $1.0\%$ on HMDB-51 and UCF-101, respectively.
Even more impressively, our approach is able to out-perform most prior works that use AudioSet~\citep{AudioSet} pre-training, which is around 10$\times$ larger than Kinetics-400.
This shows how effective and data-efficient our approach is, significantly closing the gap to supervised learning.

\paragraph{Video Action Retrieval.}

Lastly, we directly evaluate the transfer-ability of our pretrained representations on action retrieval on UCF-101 and HMDB-51.
Similarly to full fine-tuning setting, we outperform all prior works.

\setlength{\tabcolsep}{3.5pt}
\begin{table}
\centering
\footnotesize
  \begin{tabular}{lccc c  ccc}
  \toprule
       & \multicolumn{3}{c}{\textbf{UCF}} && \multicolumn{3}{c}{\textbf{HMDB}} \\
       \cmidrule{2-4} \cmidrule{6-8}
        \multicolumn{1}{r}{Recall @} & 1 & 5  & 20 && 1 & 5 & 20   \\
\midrule
    MemDPC~\cite{Han2020MemoryaugmentedDP}   & 20.2 & 40.4 & 64.7  && 7.7 & 25.7 & 57.7  \\
    VSP~\cite{cho2020selfsupervised}         & 24.6 & 41.9 & 62.7  && 10.3& 26.6 & 76.8 \\
    SeLaVi~\cite{asano2020labelling}         & 52.0 & 68.6 & 84.5  && 24.8 & 47.6 &75.5 \\
    CoCLR~\cite{Han2020SelfsupervisedCF}     & 55.9 & 70.8 & 82.5  && 26.1 & 45.8  & 69.7  \\
    GDT~\cite{Patrick2020MultimodalSF}       & 57.4 & 73.4 & 88.1  && 25.4 & 51.4  & 75.0  \\
\midrule
    \textbf{Ours:  \methodname}                                      & $\bf{59.1}$ & $\bf{76.2}$ & $\bf{88.1}$  && $\bf{26.3}$ & $\bf{49.2}$  & $\bf{76.4}$  \\
\bottomrule
  \end{tabular}
\vspace{-0.5em}
  \caption{\textbf{Comparison to SoTA for retrieval.} Nearest neighbor action retrieval performance @$k=\{1,5,20\}$.  \label{tab:ret}}
\end{table}

\section{Conclusion}
We have address two shortcomings of current self-supervised video representation learning: insufficient spatial invariance, especially compared to the image domain, and inadequate modelling of time.
We have introduced \methodname, improving spatial invariance at very little cost by implementing cropping in feature space, and improving modelling of time via a shallow transformer.
Our method brings self-supervised video representation learning one step closer to the supervised case, providing significant gains w.r.t.~the state-of-the-art.



\paragraph{Acknowledgements.}
We are grateful for support from the Rhodes Trust (M.P.), Facebook (M.P.), EPSRC Centre for Doctoral Training in Autonomous Intelligent Machines \& Systems [EP/L015897/1] (M.P. and Y.A.), the Qualcomm Fellowship (Y.A.), and the Royal Academy of Engineering under the Research Fellowship scheme (J.F.H.).
This work is also supported by the DARPA grants funded under the AIDA program (FA8750-18-2-0018) and the GAILA program (award HR00111990063) (P.H.).
We also thank Tengda Han for helpful discussions and feedback.

{\small\bibliographystyle{ieee_fullname}\bibliography{egbib,refs,vedaldi_general}}

\begin{thebibliography}{100}\itemsep=-1pt

\bibitem{afouras2018deep}
Triantafyllos Afouras, Joon~Son Chung, Andrew Senior, Oriol Vinyals, and Andrew
  Zisserman.
\newblock Deep audio-visual speech recognition.
\newblock {\em IEEE transactions on pattern analysis and machine intelligence},
  2018.

\bibitem{afouras2018conversation}
Triantafyllos Afouras, Joon~Son Chung, and Andrew Zisserman.
\newblock The conversation: Deep audio-visual speech enhancement.
\newblock {\em Interspeech}, 2018.

\bibitem{afouras2018deep_lip_reading}
Triantafyllos Afouras, Joon~Son Chung, and Andrew Zisserman.
\newblock Deep lip reading: A comparison of models and an online application.
\newblock 2018.

\bibitem{Afouras2020SelfSupervisedLO}
Triantafyllos Afouras, Andrew Owens, Joon~Son Chung, and Andrew Zisserman.
\newblock Self-supervised learning of audio-visual objects from video.
\newblock In {\em ECCV}, 2020.

\bibitem{alayrac2020selfsupervised}
Jean-Baptiste Alayrac, Adrià Recasens, Rosalia Schneider, Relja Arandjelović,
  Jason Ramapuram, Jeffrey~De Fauw, Lucas Smaira, Sander Dieleman, and Andrew
  Zisserman.
\newblock Self-supervised multimodal versatile networks.
\newblock In {\em NeurIPS}, 2020.

\bibitem{alwassel2019self}
Humam Alwassel, Bruno Korbar, Dhruv Mahajan, Lorenzo Torresani, Bernard Ghanem,
  and Du Tran.
\newblock Self-supervised learning by cross-modal audio-video clustering.
\newblock In {\em NeurIPS}, 2020.

\bibitem{Arandjelovic17}
Relja Arandjelovic and Andrew Zisserman.
\newblock Look, listen and learn.
\newblock In {\em ICCV}, 2017.

\bibitem{arandjelovic2018objects}
Relja Arandjelovi\'{c} and Andrew Zisserman.
\newblock Objects that sound.
\newblock In {\em ECCV}, 2018.

\bibitem{asano2020labelling}
Yuki~M. Asano, Mandela Patrick, Christian Rupprecht, and Andrea Vedaldi.
\newblock Labelling unlabelled videos from scratch with multi-modal
  self-supervision.
\newblock In {\em NeurIPS}, 2020.

\bibitem{asano2020a}
Yuki~M Asano, Christian Rupprecht, and Andrea Vedaldi.
\newblock A critical analysis of self-supervision, or what we can learn from a
  single image.
\newblock In {\em ICLR}, 2020.

\bibitem{asano2020self}
Yuki~M Asano, Christian Rupprecht, and Andrea Vedaldi.
\newblock Self-labelling via simultaneous clustering and representation
  learning.
\newblock In {\em ICLR}, 2020.

\bibitem{aytar2016soundnet}
Yusuf Aytar, Carl Vondrick, and Antonio Torralba.
\newblock Soundnet: Learning sound representations from unlabeled video.
\newblock In {\em NeurIPS}, 2016.

\bibitem{bachman2019learning}
Philip Bachman, R~Devon Hjelm, and William Buchwalter.
\newblock Learning representations by maximizing mutual information across
  views.
\newblock In {\em NeurIPS}, 2019.

\bibitem{benaim2020speednet}
Sagie Benaim, Ariel Ephrat, Oran Lang, Inbar Mosseri, William~T. Freeman,
  Michael Rubinstein, Michal Irani, and Tali Dekel.
\newblock Speednet: Learning the speediness in videos.
\newblock In {\em CVPR}, 2020.

\bibitem{timesformer}
Gedas Bertasius, Heng Wang, and Lorenzo Torresani.
\newblock Is space-time attention all you need for video understanding?, 2021.

\bibitem{buchler2018improving}
Uta Buchler, Biagio Brattoli, and Bjorn Ommer.
\newblock Improving spatiotemporal self-supervision by deep reinforcement
  learning.
\newblock In {\em ECCV}, 2018.

\bibitem{carion2020endtoend}
Nicolas Carion, Francisco Massa, Gabriel Synnaeve, Nicolas Usunier, Alexander
  Kirillov, and Sergey Zagoruyko.
\newblock End-to-end object detection with transformers.
\newblock In {\em ECCV}, 2020.

\bibitem{caron2018deep}
Mathilde Caron, Piotr Bojanowski, Armand Joulin, and Matthijs Douze.
\newblock Deep clustering for unsupervised learning of visual features.
\newblock In {\em ECCV}, 2018.

\bibitem{caron2019unsupervised}
Mathilde Caron, Piotr Bojanowski, Julien Mairal, and Armand Joulin.
\newblock Unsupervised pre-training of image features on non-curated data.
\newblock In {\em ICCV}, 2019.

\bibitem{caron2020swav}
Mathilde Caron, Ishan Misra, Julien Mairal, Priya Goyal, Piotr Bojanowski, and
  Armand Joulin.
\newblock Unsupervised learning of visual features by contrasting cluster
  assignments.
\newblock In {\em NeurIPS}, 2020.

\bibitem{I3D}
Joao Carreira and Andrew Zisserman.
\newblock Quo vadis, action recognition? a new model and the kinetics dataset.
\newblock In {\em CVPR}, 2017.

\bibitem{chen2019audiovisual}
Changan Chen, Unnat Jain, Carl Schissler, Sebastia Vicenc~Amengual Gari, Ziad
  Al-Halah, Vamsi~Krishna Ithapu, Philip Robinson, and Kristen Grauman.
\newblock Soundspaces: Audio-visual navigation in 3d environments.
\newblock In {\em ECCV}, 2020.

\bibitem{chen2020generative}
Mark Chen, Alec Radford, Rewon Child, Jeff Wu, Heewoo Jun, Prafulla Dhariwal,
  David Luan, and Ilya Sutskever.
\newblock Generative pretraining from pixels.
\newblock In {\em ICML}, 2020.

\bibitem{Chen2020ASF}
Ting Chen, Simon Kornblith, Mohammad Norouzi, and Geoffrey Hinton.
\newblock A simple framework for contrastive learning of visual
  representations.
\newblock In {\em ICML}, 2020.

\bibitem{cho2020selfsupervised}
Hyeon Cho, Taehoon Kim, Hyung~Jin Chang, and Wonjun Hwang.
\newblock Self-supervised spatio-temporal representation learning using
  variable playback speed prediction.
\newblock {\em arXiv preprint arXiv:2003.02692}, 2020.

\bibitem{Chung_2017_lip_reading}
Joon~Son Chung, Andrew Senior, Oriol Vinyals, and Andrew Zisserman.
\newblock Lip reading sentences in the wild.
\newblock {\em CVPR}, 2017.

\bibitem{Chung16a}
Joon~Son Chung and Andrew Zisserman.
\newblock Out of time: automated lip sync in the wild.
\newblock In {\em Workshop on Multi-view Lip-reading, ACCV}, 2016.

\bibitem{Cubuk2019RandAugmentPD}
E. Cubuk, Barret Zoph, Jonathon Shlens, and Quoc~V. Le.
\newblock Randaugment: Practical data augmentation with no separate search.
\newblock In {\em CVPRW}, 2020.

\bibitem{autoaugment}
Ekin~Dogus Cubuk, Barret Zoph, Dandelion Mane, Vijay Vasudevan, and Quoc~V. Le.
\newblock Autoaugment: Learning augmentation policies from data.
\newblock In {\em CVPR}, 2020.

\bibitem{dalal05histogram}
Navneet Dalal and Bill Triggs.
\newblock Histograms of oriented gradients for human detection.
\newblock In {\em Proc. {CVPR}}, 2005.

\bibitem{deng2009imagenet}
Jia Deng, Wei Dong, Richard Socher, Li-Jia Li, Kai Li, and Li Fei-Fei.
\newblock Imagenet: A large-scale hierarchical image database.
\newblock In {\em CVPR}, 2009.

\bibitem{desai2020virtex}
Karan Desai and Justin Johnson.
\newblock Virtex: Learning visual representations from textual annotations,
  2020.

\bibitem{devlin18bert:}
Jacob Devlin, Ming{-}Wei Chang, Kenton Lee, and Kristina Toutanova.
\newblock {BERT:} pre-training of deep bidirectional transformers for language
  understanding.
\newblock {\em CoRR}, abs/1810.04805, 2018.

\bibitem{Diba_2019}
Ali Diba, Vivek Sharma, Luc~Van Gool, and Rainer Stiefelhagen.
\newblock Dynamonet: Dynamic action and motion network.
\newblock In {\em ICCV}, 2019.

\bibitem{doersch2015unsupervised}
Carl Doersch, Abhinav Gupta, and Alexei~A Efros.
\newblock Unsupervised visual representation learning by context prediction.
\newblock In {\em ICCV}, 2015.

\bibitem{doersch2020crosstransformers}
Carl Doersch, Ankush Gupta, and Andrew Zisserman.
\newblock Crosstransformers: spatially-aware few-shot transfer.
\newblock In {\em NeurIPS}, 2020.

\bibitem{donahue2016longterm}
Jeff Donahue, Lisa~Anne Hendricks, Marcus Rohrbach, Subhashini Venugopalan,
  Sergio Guadarrama, Kate Saenko, and Trevor Darrell.
\newblock Long-term recurrent convolutional networks for visual recognition and
  description.
\newblock In {\em CVPR}, 2015.

\bibitem{dosovitskiy2020image}
Alexey Dosovitskiy, Lucas Beyer, Alexander Kolesnikov, Dirk Weissenborn,
  Xiaohua Zhai, Thomas Unterthiner, Mostafa Dehghani, Matthias Minderer, Georg
  Heigold, Sylvain Gelly, Jakob Uszkoreit, and Neil Houlsby.
\newblock An image is worth 16x16 words: Transformers for image recognition at
  scale, 2020.

\bibitem{dosovitskiy2015discriminative}
Alexey Dosovitskiy, Philipp Fischer, Jost~Tobias Springenberg, Martin
  Riedmiller, and Thomas Brox.
\newblock Discriminative unsupervised feature learning with exemplar
  convolutional neural networks.
\newblock {\em TPAMI}, 38(9), 2015.

\bibitem{Epstein2020OopsPU}
D. Epstein, Boyuan Chen, and Carl Vondrick.
\newblock Oops! predicting unintentional action in video.
\newblock In {\em CVPR}, 2020.

\bibitem{slowfast}
Christoph Feichtenhofer, Haoqi Fan, Jitendra Malik, and Kaiming He.
\newblock Slowfast networks for video recognition.
\newblock In {\em ICCV}, 2019.

\bibitem{gabeur2020multimodal}
Valentin Gabeur, Chen Sun, Karteek Alahari, and Cordelia Schmid.
\newblock Multi-modal transformer for video retrieval.
\newblock In {\em ECCV}, 2020.

\bibitem{gao2019listen}
Ruohan Gao, Tae-Hyun Oh, Kristen Grauman, and Lorenzo Torresani.
\newblock Listen to look: Action recognition by previewing audio.
\newblock In {\em CVPR}, 2020.

\bibitem{AudioSet}
Jort~F. Gemmeke, Daniel P.~W. Ellis, Dylan Freedman, Aren Jansen, Wade
  Lawrence, R.~Channing Moore, Manoj Plakal, and Marvin Ritter.
\newblock Audio set: An ontology and human-labeled dataset for audio events.
\newblock In {\em ICASSP}, 2017.

\bibitem{Ghadiyaram2019}
Deepti Ghadiyaram, Du Tran, and Dhruv Mahajan.
\newblock {Large-scale weakly-supervised pre-training for video action
  recognition}.
\newblock In {\em CVPR}, 2019.

\bibitem{gidaris2018unsupervised}
Spyros Gidaris, Praveer Singh, and Nikos Komodakis.
\newblock Unsupervised representation learning by predicting image rotations.
\newblock {\em ICLR}, 2018.

\bibitem{girdhar2019video}
Rohit Girdhar, João Carreira, Carl Doersch, and Andrew Zisserman.
\newblock Video action transformer network.
\newblock In {\em CVPR}, 2019.

\bibitem{Girdhar17}
Rohit Girdhar, Deva Ramanan, Abhinav Gupta, Josef Sivic, and Bryan~C. Russell.
\newblock Actionvlad: Learning spatio-temporal aggregation for action
  classification.
\newblock In {\em CVPR}, 2017.

\bibitem{grill2020bootstrap}
Jean-Bastien Grill, Florian Strub, Florent Altché, Corentin Tallec, Pierre~H.
  Richemond, Elena Buchatskaya, Carl Doersch, Bernardo~Avila Pires,
  Zhaohan~Daniel Guo, Mohammad~Gheshlaghi Azar, Bilal Piot, Koray Kavukcuoglu,
  Rémi Munos, and Michal Valko.
\newblock Bootstrap your own latent: A new approach to self-supervised
  learning.
\newblock In {\em NeurIPS}, 2020.

\bibitem{gutmann2010noise}
Michael Gutmann and Aapo Hyv{\"a}rinen.
\newblock Noise-contrastive estimation: A new estimation principle for
  unnormalized statistical models.
\newblock In {\em AISTATS}, 2010.

\bibitem{Hadsell2006DimensionalityRB}
Raia Hadsell, Sumit Chopra, and Yann LeCun.
\newblock Dimensionality reduction by learning an invariant mapping.
\newblock In {\em CVPR}, 2006.

\bibitem{han2019video}
Tengda Han, Weidi Xie, and Andrew Zisserman.
\newblock Video representation learning by dense predictive coding.
\newblock In {\em ICCVW}, 2019.

\bibitem{Han2020MemoryaugmentedDP}
Tengda Han, Weidi Xie, and Andrew Zisserman.
\newblock Memory-augmented dense predictive coding for video representation
  learning.
\newblock In {\em ECCV}, 2020.

\bibitem{Han2020SelfsupervisedCF}
Tengda Han, Weidi Xie, and Andrew Zisserman.
\newblock Self-supervised co-training for video representation learning.
\newblock In {\em NeurIPS}, 2020.

\bibitem{hara2017learning}
Kensho Hara, Hirokatsu Kataoka, and Yutaka Satoh.
\newblock Learning spatio-temporal features with 3d residual networks for
  action recognition.
\newblock In {\em ICCVW}, 2017.

\bibitem{harwath2018jointly}
David Harwath, Adria Recasens, D{\'\i}dac Sur{\'\i}s, Galen Chuang, Antonio
  Torralba, and James Glass.
\newblock Jointly discovering visual objects and spoken words from raw sensory
  input.
\newblock In {\em ECCV}, 2018.

\bibitem{he2019momentum}
Kaiming He, Haoqi Fan, Yuxin Wu, Saining Xie, and Ross Girshick.
\newblock Momentum contrast for unsupervised visual representation learning.
\newblock In {\em CVPR}, 2020.

\bibitem{he19momentum}
Kaiming He, Haoqi Fan, Yuxin Wu, Saining Xie, and Ross~B. Girshick.
\newblock Momentum contrast for unsupervised visual representation learning.
\newblock {\em arXiv.cs}, abs/1911.05722, 2019.

\bibitem{he16resnet}
Kaiming He, Xiangyu Zhang, Shaoqing Ren, and Jian Sun.
\newblock {Deep Residual Learning for Image Recognition}.
\newblock In {\em {CVPR}}, 2016.

\bibitem{hu2018deep}
Di Hu, Feiping Nie, and Xuelong Li.
\newblock Deep multimodal clustering for unsupervised audiovisual learning.
\newblock In {\em CVPR}, 2019.

\bibitem{hussein2019timeception}
Noureldien Hussein, Efstratios Gavves, and Arnold W.~M. Smeulders.
\newblock Timeception for complex action recognition.
\newblock In {\em CVPR}, 2019.

\bibitem{bert18}
Kenton~Lee Jacob~Devlin, Ming-Wei~Chang and Kristina Toutanova.
\newblock {BERT}: Pre-training of deep bidirectional transformers for language
  understanding.
\newblock In {\em NAACL}, 2018.

\bibitem{Jenni2020VideoRL}
S. Jenni, Givi Meishvili, and P. Favaro.
\newblock Learning video representations by transforming time.
\newblock In {\em ECCV}, 2020.

\bibitem{ji2018invariant}
Xu Ji, João~F. Henriques, and Andrea Vedaldi.
\newblock Invariant information clustering for unsupervised image
  classification and segmentation.
\newblock In {\em ICCV}, 2019.

\bibitem{jing2018self}
Longlong Jing and Yingli Tian.
\newblock Self-supervised spatiotemporal feature learning by video geometric
  transformations.
\newblock {\em arXiv preprint arXiv:1811.11387}, 2018.

\bibitem{kalantidis2020hard}
Yannis Kalantidis, Mert~Bulent Sariyildiz, Noe Pion, Philippe Weinzaepfel, and
  Diane Larlus.
\newblock Hard negative mixing for contrastive learning.
\newblock In {\em NeurIPS}, 2020.

\bibitem{kant2020spatially}
Yash Kant, Dhruv Batra, Peter Anderson, Alex Schwing, Devi Parikh, Jiasen Lu,
  and Harsh Agrawal.
\newblock Spatially aware multimodal transformers for textvqa.
\newblock In {\em ECCV}, 2020.

\bibitem{Sports1M}
Andrej Karpathy, George Toderici, Sanketh Shetty, Thomas Leung, Rahul
  Sukthankar, and Li Fei-Fei.
\newblock Large-scale video classification with convolutional neural networks.
\newblock In {\em CVPR}, 2014.

\bibitem{kinetics}
Will Kay, Joao Carreira, Karen Simonyan, Brian Zhang, Chloe Hillier, Sudheendra
  Vijayanarasimhan, Fabio Viola, Tim Green, Trevor Back, Paul Natsev, et~al.
\newblock The kinetics human action video dataset.
\newblock {\em arXiv preprint arXiv:1705.06950}, 2017.

\bibitem{kazakos2019epicfusion}
Evangelos Kazakos, Arsha Nagrani, Andrew Zisserman, and Dima Damen.
\newblock Epic-fusion: Audio-visual temporal binding for egocentric action
  recognition.
\newblock In {\em ICCV}, 2019.

\bibitem{kim2019self}
Dahun Kim, Donghyeon Cho, and In~So Kweon.
\newblock Self-supervised video representation learning with space-time cubic
  puzzles.
\newblock In {\em AAAI}, 2019.

\bibitem{kingma15adam}
Diederik~P. Kingma and Jimmy Ba.
\newblock Adam: A method for stochastic optimization.
\newblock In {\em ICLR}, 2015.

\bibitem{korbar2020video}
Bruno Korbar, Fabio Petroni, Rohit Girdhar, and Lorenzo Torresani.
\newblock Video understanding as machine translation, 2020.

\bibitem{avts}
Bruno Korbar, Du Tran, and Lorenzo Torresani.
\newblock Cooperative learning of audio and video models from self-supervised
  synchronization.
\newblock In {\em NeurIPS}, 2018.

\bibitem{korbar2019scsampler}
Bruno Korbar, Du Tran, and Lorenzo Torresani.
\newblock Scsampler: Sampling salient clips from video for efficient action
  recognition.
\newblock In {\em ICCV}, 2019.

\bibitem{Krizhevsky12}
Alex Krizhevsky, Ilya Sutskever, and Geoffrey~E. Hinton.
\newblock Imagenet classification with deep convolutional neural networks.
\newblock In {\em NeurIPS}, 2012.

\bibitem{HMDB51}
H. Kuehne, H. Jhuang, E. Garrote, T. Poggio, and T. Serre.
\newblock {HMDB}: a large video database for human motion recognition.
\newblock In {\em ICCV}, 2011.

\bibitem{kuo2020featmatch}
Chia-Wen Kuo, Chih-Yao Ma, Jia-Bin Huang, and Zsolt Kira.
\newblock Featmatch: Feature-based augmentation for semi-supervised learning.
\newblock In {\em ECCV}, 2020.

\bibitem{lee2017unsupervised}
Hsin-Ying Lee, Jia-Bin Huang, Maneesh Singh, and Ming-Hsuan Yang.
\newblock Unsupervised representation learning by sorting sequences.
\newblock In {\em ICCV}, 2017.

\bibitem{lee2021parameter}
Sangho Lee, Youngjae Yu, Gunhee Kim, Thomas Breuel, Jan Kautz, and Yale Song.
\newblock Parameter efficient multimodal transformers for video representation
  learning.
\newblock In {\em International Conference on Learning Representations}, 2021.

\bibitem{lewis2019bart}
Mike Lewis, Yinhan Liu, Naman Goyal, Marjan Ghazvininejad, Abdelrahman Mohamed,
  Omer Levy, Ves Stoyanov, and Luke Zettlemoyer.
\newblock Bart: Denoising sequence-to-sequence pre-training for natural
  language generation, translation, and comprehension.
\newblock In {\em ACL}, 2020.

\bibitem{li2020centerwise}
Hao Li, Xiaopeng Zhang, Ruoyu Sun, Hongkai Xiong, and Qi Tian.
\newblock Center-wise local image mixture for contrastive representation
  learning, 2020.

\bibitem{li2020prototypical}
Junnan Li, Pan Zhou, Caiming Xiong, Richard Socher, and Steven~CH Hoi.
\newblock Prototypical contrastive learning of unsupervised representations.
\newblock {\em arXiv preprint arXiv:2005.04966}, 2020.

\bibitem{li2019visualbert}
Liunian~Harold Li, Mark Yatskar, Da Yin, Cho-Jui Hsieh, and Kai-Wei Chang.
\newblock Visualbert: A simple and performant baseline for vision and language,
  2019.

\bibitem{li2020learning}
Tianhao Li and Limin Wang.
\newblock Learning spatiotemporal features via video and text pair
  discrimination.
\newblock {\em arXiv preprint arXiv:2001.05691}, 2020.

\bibitem{locatello2020objectcentric}
Francesco Locatello, Dirk Weissenborn, Thomas Unterthiner, Aravindh Mahendran,
  Georg Heigold, Jakob Uszkoreit, Alexey Dosovitskiy, and Thomas Kipf.
\newblock Object-centric learning with slot attention.
\newblock In {\em NeurIPS}, 2020.

\bibitem{lowe99object}
David~G. Lowe.
\newblock Object recognition from local scale-invariant features.
\newblock In {\em Proc. {ICCV}}, 1999.

\bibitem{lu2019vilbert}
Jiasen Lu, Dhruv Batra, Devi Parikh, and Stefan Lee.
\newblock Vilbert: Pretraining task-agnostic visiolinguistic representations
  for vision-and-language tasks.
\newblock In {\em NeurIPS}, 2019.

\bibitem{luo2017understanding}
Wenjie Luo, Yujia Li, Raquel Urtasun, and Richard Zemel.
\newblock Understanding the effective receptive field in deep convolutional
  neural networks, 2017.

\bibitem{ma2020learning}
Shuang Ma, Zhaoyang Zeng, Daniel McDuff, and Yale Song.
\newblock Learning audio-visual representations with active contrastive coding,
  2020.

\bibitem{ma2021contrastive}
Shuang Ma, Zhaoyang Zeng, Daniel McDuff, and Yale Song.
\newblock Contrastive self-supervised learning of global-local audio-visual
  representations, 2021.

\bibitem{miech17learning}
Antoine Miech, Jean{-}Baptiste Alayrac, Piotr Bojanowski, Ivan Laptev, and
  Josef Sivic.
\newblock Learning from video and text via large-scale discriminative
  clustering.
\newblock In {\em Proc. {ICCV}}, 2017.

\bibitem{miech2019endtoend}
Antoine Miech, Jean-Baptiste Alayrac, Lucas Smaira, Ivan Laptev, Josef Sivic,
  and Andrew Zisserman.
\newblock End-to-end learning of visual representations from uncurated
  instructional videos.
\newblock In {\em CVPR}, 2020.

\bibitem{miech19howto100m}
Antoine Miech, Dimitri Zhukov, Jean-Baptiste Alayrac, Makarand Tapaswi, Ivan
  Laptev, and Josef Sivic.
\newblock Howto100{M}: Learning a text-video embedding by watching hundred
  million narrated video clips.
\newblock In {\em ICCV}, 2019.

\bibitem{misra2019selfsupervised}
Ishan Misra and Laurens van~der Maaten.
\newblock Self-supervised learning of pretext-invariant representations.
\newblock In {\em CVPR}, 2020.

\bibitem{misra2016shuffle}
Ishan Misra, C~Lawrence Zitnick, and Martial Hebert.
\newblock Shuffle and learn: unsupervised learning using temporal order
  verification.
\newblock In {\em ECCV}, 2016.

\bibitem{morgado2020avid}
Pedro Morgado, Nuno Vasconcelos, and Ishan Misra.
\newblock Audio-visual instance discrimination with cross-modal agreement.
\newblock {\em arXiv preprint arXiv:2004.12943}, 2020.

\bibitem{nagrani2020}
Arsha Nagrani, Chen Sun, David Ross, Rahul Sukthankar, Cordelia Schmid, and
  Andrew Zisserman.
\newblock Speech2action: Cross-modal supervision for action recognition.
\newblock In {\em CVPR}, 2020.

\bibitem{neimark2021VTN}
Daniel Neimark, Omri Bar, Maya Zohar, and Dotan Asselmann.
\newblock Video transformer network, 2021.

\bibitem{noroozi2016unsupervised}
Mehdi Noroozi and Paolo Favaro.
\newblock Unsupervised learning of visual representations by solving jigsaw
  puzzles.
\newblock In {\em ECCV}, 2016.

\bibitem{oord2018representation}
Aaron van~den Oord, Yazhe Li, and Oriol Vinyals.
\newblock Representation learning with contrastive predictive coding.
\newblock {\em arXiv preprint arXiv:1807.03748}, 2018.

\bibitem{owens2018audio}
Andrew Owens and Alexei~A Efros.
\newblock Audio-visual scene analysis with self-supervised multisensory
  features.
\newblock In {\em ECCV}, 2018.

\bibitem{Park2019SpecAugmentAS}
D. Park, William Chan, Y. Zhang, Chung-Cheng Chiu, Barret Zoph, E.~D. Cubuk,
  and Quoc~V. Le.
\newblock Specaugment: A simple data augmentation method for automatic speech
  recognition.
\newblock In {\em INTERSPEECH}, 2019.

\bibitem{parmar2018image}
Niki Parmar, Ashish Vaswani, Jakob Uszkoreit, Łukasz Kaiser, Noam Shazeer,
  Alexander Ku, and Dustin Tran.
\newblock Image transformer.
\newblock In {\em ICML}, 2018.

\bibitem{pathak2016context}
Deepak Pathak, Philipp Krahenbuhl, Jeff Donahue, Trevor Darrell, and Alexei~A
  Efros.
\newblock Context encoders: Feature learning by inpainting.
\newblock In {\em CVPR}, 2016.

\bibitem{Patrick2020MultimodalSF}
Mandela Patrick, Yuki~Markus Asano, Ruth Fong, Jo{\~a}o~F. Henriques, G. Zweig,
  and A. Vedaldi.
\newblock Multi-modal self-supervision from generalized data transformations.
\newblock {\em ArXiv}, abs/2003.04298, 2020.

\bibitem{patrick2020supportset}
Mandela Patrick, Po-Yao Huang, Yuki Asano, Florian Metze, Alexander Hauptmann,
  João Henriques, and Andrea Vedaldi.
\newblock Support-set bottlenecks for video-text representation learning, 2020.

\bibitem{Piczak2015}
Karol~J. Piczak.
\newblock Environmental sound classification with convolutional neural
  networks.
\newblock {\em MLSP}, 2015.

\bibitem{esc50}
Karol~J. Piczak.
\newblock Esc: Dataset for environmental sound classification.
\newblock In {\em ACM Multimedia}, 2015.

\bibitem{piergiovanni2020evolving}
AJ Piergiovanni, Anelia Angelova, and Michael~S. Ryoo.
\newblock Evolving losses for unsupervised video representation learning.
\newblock In {\em CVPR}, 2020.

\bibitem{potamianos2003recent}
Gerasimos Potamianos, Chalapathy Neti, Guillaume Gravier, Ashutosh Garg, and
  Andrew~W Senior.
\newblock Recent advances in the automatic recognition of audiovisual speech.
\newblock {\em Proceedings of the IEEE}, 91(9):1306--1326, 2003.

\bibitem{Qian2020SpatiotemporalCV}
Rui Qian, Tianjian Meng, Boqing Gong, Ming-Hsuan Yang, H. Wang, Serge~J.
  Belongie, and Yin Cui.
\newblock Spatiotemporal contrastive video representation learning.
\newblock 2020.

\bibitem{radford2019language}
Alec Radford, Jeffrey Wu, Rewon Child, David Luan, Dario Amodei, and Ilya
  Sutskever.
\newblock Language models are unsupervised multitask learners.
\newblock {\em OpenAI Blog}, 1(8):9, 2019.

\bibitem{raffel2019exploring}
Colin Raffel, Noam Shazeer, Adam Roberts, Katherine Lee, Sharan Narang, Michael
  Matena, Yanqi Zhou, Wei Li, and Peter~J Liu.
\newblock Exploring the limits of transfer learning with a unified text-to-text
  transformer.
\newblock {\em arXiv preprint arXiv:1910.10683}, 2019.

\bibitem{ren2015faster}
Shaoqing Ren, Kaiming He, Ross Girshick, and Jian Sun.
\newblock {Faster R-CNN}: Towards real-time object detection with region
  proposal networks.
\newblock In {\em NeurIPS}, 2015.

\bibitem{rnh2013}
Guido Roma, Waldo Nogueira, and Perfecto Herrera.
\newblock Recurrence quantification analysis features for environmental sound
  recognition.
\newblock {\em WASPAA}, 2013.

\bibitem{sailor2017}
Hardik~B. Sailor, Dharmesh~M Agrawal, and Hemant~A Patil.
\newblock Unsupervised filterbank learning using convolutional restricted
  boltzmann machine for environmental sound classification.
\newblock In {\em INTERSPEECH}, 2017.

\bibitem{sariyildiz2020learning}
Mert~Bulent Sariyildiz, Julien Perez, and Diane Larlus.
\newblock Learning visual representations with caption annotations.
\newblock In {\em ECCV}, 2020.

\bibitem{simonyan2014twostream}
Karen Simonyan and Andrew Zisserman.
\newblock Two-stream convolutional networks for action recognition in videos.
\newblock In {\em NeurIPS}, 2014.

\bibitem{UCF101}
Khurram Soomro, Amir~Roshan Zamir, and Mubarak Shah.
\newblock {UCF101}: A dataset of 101 human action classes from videos in the
  wild.
\newblock In {\em CRCV-TR-12-01}, 2012.

\bibitem{stowell2015}
Dan Stowell, Dimitrios Giannoulis, Emmanouil Benetos, Mathieu Lagrange, and
  Mark~D. Plumbley.
\newblock Detection and classification of acoustic scenes and events.
\newblock {\em TM}, 2015.

\bibitem{DCASE}
D. {Stowell}, D. {Giannoulis}, E. {Benetos}, M. {Lagrange}, and M.~D.
  {Plumbley}.
\newblock Detection and classification of acoustic scenes and events.
\newblock {\em IEEE Transactions on Multimedia}, 2015.

\bibitem{Stroud2020LearningVR}
Jonathan~C. Stroud, D. Ross, Chen Sun, Jun Deng, R. Sukthankar, and C. Schmid.
\newblock Learning video representations from textual web supervision.
\newblock {\em ArXiv}, abs/2007.14937, 2020.

\bibitem{su2019vlbert}
Weijie Su, Xizhou Zhu, Yue Cao, Bin Li, Lewei Lu, Furu Wei, and Jifeng Dai.
\newblock Vl-bert: Pre-training of generic visual-linguistic representations.
\newblock In {\em ICLR}, 2020.

\bibitem{sun2019contrastive}
Chen Sun, Fabien Baradel, Kevin Murphy, and Cordelia Schmid.
\newblock Contrastive bidirectional transformer for temporal representation
  learning.
\newblock {\em arXiv preprint arXiv:1906.05743}, 2019.

\bibitem{sun2019videobert}
Chen Sun, Austin Myers, Carl Vondrick, Kevin Murphy, and Cordelia Schmid.
\newblock Videobert: A joint model for video and language representation
  learning.
\newblock In {\em ICCV}, 2019.

\bibitem{Tan_2019}
Hao Tan and Mohit Bansal.
\newblock Lxmert: Learning cross-modality encoder representations from
  transformers.
\newblock In {\em EMNLP}, 2019.

\bibitem{tan2020vokenization}
Hao Tan and Mohit Bansal.
\newblock Vokenization: Improving language understanding with contextualized,
  visual-grounded supervision.
\newblock In {\em EMNLP}, 2020.

\bibitem{terrance2017dataset}
V Terrance and W~Taylor Graham.
\newblock Dataset augmentation in feature space.
\newblock In {\em Proceedings of the international conference on machine
  learning (ICML), workshop track}, 2017.

\bibitem{tian2019contrastive}
Yonglong Tian, Dilip Krishnan, and Phillip Isola.
\newblock Contrastive multiview coding.
\newblock In {\em ECCV}, 2020.

\bibitem{tian2020makes}
Yonglong Tian, Chen Sun, Ben Poole, Dilip Krishnan, Cordelia Schmid, and
  Phillip Isola.
\newblock What makes for good views for contrastive learning.
\newblock In {\em NeurIPS}, 2020.

\bibitem{Tran15}
Du Tran, Lubomir Bourdev, Rob Fergus, Lorenzo Torresani, and Manohar Paluri.
\newblock Learning spatiotemporal features with 3d convolutional networks.
\newblock In {\em ICCV}, 2015.

\bibitem{Tran18}
Du Tran, Heng Wang, Lorenzo Torresani, Jamie Ray, Yann LeCun, and Manohar
  Paluri.
\newblock A closer look at spatiotemporal convolutions for action recognition.
\newblock In {\em CVPR}, 2018.

\bibitem{varol17}
G{\"u}l Varol, Ivan Laptev, and Cordelia Schmid.
\newblock {Long-term Temporal Convolutions for Action Recognition}.
\newblock {\em IEEE Transactions on Pattern Analysis and Machine Intelligence},
  2017.

\bibitem{vaswani2017attention}
Ashish Vaswani, Noam Shazeer, Niki Parmar, Jakob Uszkoreit, Llion Jones,
  Aidan~N Gomez, {\L}ukasz Kaiser, and Illia Polosukhin.
\newblock Attention is all you need.
\newblock In {\em NeurIPS}, 2017.

\bibitem{motion_statistics}
Jiangliu Wang, Jianbo Jiao, Linchao Bao, Shengfeng He, Yunhui Liu, and Wei Liu.
\newblock Self-supervised spatio-temporal representation learning for videos by
  predicting motion and appearance statistics.
\newblock In {\em CVPR}, 2019.

\bibitem{Wang2020SelfsupervisedVR}
Jiangliu Wang, Jianbo Jiao, and Y. Liu.
\newblock Self-supervised video representation learning by pace prediction.
\newblock In {\em ECCV}, 2020.

\bibitem{wang16temporal}
Limin Wang, Yuanjun Xiong, Yu Qiao, Dahua Lin, Xiaoou Tang, and Luc Van~Gool.
\newblock Temporal segment networks: Towards good practices for deep action
  recognition.
\newblock In {\em ECCV}, 2016.

\bibitem{XiaolongWang18}
Xiaolong Wang, Ross Girshick, Abhinav Gupta, and Kaiming He.
\newblock Non-local neural networks.
\newblock In {\em CVPR}, 2018.

\bibitem{wei2018learning}
Donglai Wei, Joseph~J Lim, Andrew Zisserman, and William~T Freeman.
\newblock Learning and using the arrow of time.
\newblock In {\em CVPR}, 2018.

\bibitem{wu2020visual}
Bichen Wu, Chenfeng Xu, Xiaoliang Dai, Alvin Wan, Peizhao Zhang, Masayoshi
  Tomizuka, Kurt Keutzer, and Peter Vajda.
\newblock Visual transformers: Token-based image representation and processing
  for computer vision, 2020.

\bibitem{Wu_2018_CVPR}
Zhirong Wu, Yuanjun Xiong, Stella~X. Yu, and Dahua Lin.
\newblock Unsupervised feature learning via non-parametric instance
  discrimination.
\newblock In {\em CVPR}, 2018.

\bibitem{xiao2020audiovisual}
Fanyi Xiao, Yong~Jae Lee, Kristen Grauman, Jitendra Malik, and Christoph
  Feichtenhofer.
\newblock Audiovisual slowfast networks for video recognition.
\newblock {\em arXiv preprint arXiv:2001.08740}, 2020.

\bibitem{xie2017rethinking}
Saining Xie, Chen Sun, Jonathan Huang, Zhuowen Tu, and Kevin Murphy.
\newblock Rethinking spatiotemporal feature learning for video understanding.
\newblock In {\em ECCV}, 2018.

\bibitem{clip_order}
Dejing Xu, Jun Xiao, Zhou Zhao, Jian Shao, Di Xie, and Yueting Zhuang.
\newblock Self-supervised spatiotemporal learning via video clip order
  prediction.
\newblock In {\em CVPR}, 2019.

\bibitem{yun2019cutmix}
Sangdoo Yun, Dongyoon Han, Seong~Joon Oh, Sanghyuk Chun, Junsuk Choe, and
  Youngjoon Yoo.
\newblock Cutmix: Regularization strategy to train strong classifiers with
  localizable features.
\newblock In {\em ICCV}, 2019.

\bibitem{zhang2020adaptive}
Jingzhao Zhang, Sai~Praneeth Karimireddy, Andreas Veit, Seungyeon Kim,
  Sashank~J Reddi, Sanjiv Kumar, and Suvrit Sra.
\newblock Why are adaptive methods good for attention models?, 2020.

\bibitem{zhang2016colorful}
Richard Zhang, Phillip Isola, and Alexei~A Efros.
\newblock Colorful image colorization.
\newblock In {\em ECCV}, 2016.

\bibitem{zhang2017split}
Richard Zhang, Phillip Isola, and Alexei~A Efros.
\newblock Split-brain autoencoders: Unsupervised learning by cross-channel
  prediction.
\newblock In {\em CVPR}, 2017.

\bibitem{zhao2019sound}
Hang Zhao, Chuang Gan, Wei-Chiu Ma, and Antonio Torralba.
\newblock The sound of motions.
\newblock In {\em ICCV}, 2019.

\bibitem{zhao2018sound}
Hang Zhao, Chuang Gan, Andrew Rouditchenko, Carl Vondrick, Josh McDermott, and
  Antonio Torralba.
\newblock The sound of pixels.
\newblock In {\em ECCV}, 2018.

\bibitem{abs08496}
Bolei Zhou, Alex Andonian, and Antonio Torralba.
\newblock Temporal relational reasoning in videos.
\newblock In {\em ECCV}, 2018.

\bibitem{ZhouKLOT14}
Bolei Zhou, Aditya Khosla, Àgata Lapedriza, Aude Oliva, and Antonio Torralba.
\newblock Object detectors emerge in deep scene cnns.
\newblock In {\em ICLR}, 2015.

\bibitem{zhu2020actbert}
Linchao Zhu and Yi Yang.
\newblock Actbert: Learning global-local video-text representations.
\newblock In {\em CVPR}, 2020.

\end{thebibliography}

\clearpage\newpage\section{Appendix}

\subsection{Implementation Details}
While videos in Kinetics are 10 seconds long, we randomly sample either 1-second (30 frames), or 2-second (60 frames) clips from the 30fps videos.
For the R(2+1)-D-18 visual encoder, the dimensions of the $res5$ feature map before spatial pooling is $512$ x $T$ x $7$ x $7$ for a $112$ x $112$ resolution video, where $T = 4$ for 30-frame (1 second) input, and $T = 8$ for 60-frame (2 second) input. 
After spatial pooling, we use either average pooling or a transformer as the temporal pooling function for the visual encoder, but always use average pooling for the audio encoder. 
The transformer's layers dimensionality are set to 512-D.
Both encoders produce a fixed-dimensional representation vectors after temporal aggregation (512-D).
Both vectors are then passed through two fully-connected layers with intermediate size of $512$ to produce 256-D embedding vectors $\bz$ as in~\cite{Patrick2020MultimodalSF}.
We use these embeddings in our loss eq.~\eqref{loss:total} and train our model for $100$ epochs.
For the visual component of the video, we use a $30$ frame RGB clip as input, at $30$ fps covering 1 second.
The video clip has a spatial resolution of $112 \! \times \!112$ pixels.
For input data augmentation, we apply random crops, horizontal flips, Gaussian blur and color jittering, all clip-wise consistent, following the protocol of SimCLR~\cite{Chen2020ASF}, and we ablate multiple settings for spatial and temporal feature cropping sizes.
For the audio input, we extract a $1$-second log-mel spectrogram of dimension $257\times 199$ starting at the same time as the visual component.
We also apply volume jittering to increase the robustness of our audio features. 
We optimize this model using SGD with momentum $0.9$, weight decay $10^{-5}$ and learning rate $0.64$, with a warm-up period of $10$ epochs.
For NCE contrastive learning, the temperature $\tau$ is set as $0.1$ for cross-modal loss, and $0.5$ for the within-modal loss.
We use a mini-batch size of $8$ on each of our $64$ GPUs giving an effective batch size of $512$ for distributed training.
In our ablations, we evaluate the learned representation by finetuning the visual encoder on fold 1 of the HMDB-51~\cite{HMDB51} action recognition dataset.

\subsubsection{State-of-the-Art Experiment Details}
For our state-of-the-art model, we train for 100 epochs, using R(2+1)-D-18 visual encoder with transformer temporal attention pooling, and Resnet-9 for audio encoder. 
We use 60 frames as input, and feature-crop augmentation (space: $2\!\times \!6^2 \!+ \!4\!\times \!4^2$  \& time: $2\!\times \!3 \! + \! 1\!\times \!2$).


\subsection{Transformer Architecture Details}
We use a 2-layer transformer, with 4 attention heads, and hidden dimension 512.
The input to the transformer is the spatially averaged output of the last convolutional layer of R(2+1)D-18 video backbone. 
The transformer contextualizes features across time to output a fixed feature length representation of dimension 512, which is then passed to MLP head for contrastive learning. While transformers generally benefit from being optimized with Adam~\citep{zhang2020adaptive}, we adhere to using SGD for simplicity. 
We also do not observe any stability issues, likely because the transformer is quite shallow.


\section{Additional experiments}

\subsection{Audio Classification}
\begin{table}
\begin{tabular}{l l cc} 
\toprule
\textbf{Method}                         & \textbf{Pretraining} &\multicolumn{2}{c}{\underline{\textbf{Acc\%}}} \\
                                        &           & DCASE  & ESC50 \\
\midrule
Autoencoder \cite{aytar2016soundnet}    & -    & -        &   39.9        \\
Random Forest \cite{esc50}              & -    & -        &   44.3        \\
Piczak ConvNet \cite{Piczak2015}        & -    & -        &   64.5        \\
RNH \cite{rnh2013}                      & -    & 72       &   -           \\
Ensemble \cite{stowell2015}             & -    & 77       &   -           \\
ConvRBM \cite{sailor2017}               & -    & -        & 86.5          \\
\midrule
AVTS \cite{avts}                        & K400   & 91       & 76.7          \\
XDC \cite{alwassel2019self}             & K400   & --       & 78.0          \\
AVID \cite{morgado2020avid}             & K400   & \ul{93}  & 79.1         \\
ACC \cite{ma2020learning}               & K400   & --       & \ul{79.2}       \\
\midrule
\textbf{Ours: \methodname}                    & K400  & \bf{94}   &   \bf{81.1}           \\
\midrule
SoundNet \cite{aytar2016soundnet}       & SNet  & 88        & 74.2          \\
L3-Net \cite{Arandjelovic17}            & SNet  & 93        & 79.3          \\
AVTS \cite{avts}                        & SNet  & 94        & 82.3          \\
DMC \cite{hu2018deep}                   & SNet & --         & 82.6          \\
\midrule
AVTS \cite{avts}                        & AS   & 93        & 80.6          \\
XDC \cite{alwassel2019self}             & AS   & --        & 85.8          \\
MMV \cite{alayrac2020selfsupervised}    & AS   & --        & 86.1       \\
AVID \cite{morgado2020avid}             & AS   & \ul{96}   & \ul{89.2}      \\
GDT \cite{Patrick2020MultimodalSF}      & AS   & \bf{98}   & 88.5       \\
ACC \cite{ma2020learning}               & AS   & --        & \bf{90.8}       \\
\midrule
Human \cite{esc50}                      & -- & --      & 81.3 \\
\bottomrule
\end{tabular} 
\caption{\textbf{Audio classification.} 
Downstream task accuracies on standard audio classification benchmarks on DCASE2014 and ESC50. Dataset abbreviations
  \textbf{\ul{A}}udio\textbf{\ul{S}}et, 
  \textbf{\ul{K}}inetics\textbf{\ul{400}},
  \textbf{\ul{S}}ound\textbf{\ul{Net}},
  \label{tab:audio}}
\end{table}
For completeness, we also present audio classification results on ESC-50 [108] and DCASE-2014~\citep{DCASE}. 
ESC- 50~\citep{esc50} is an environmental sound classification dataset which has 2K sound clips of 50 different audio classes. 
ESC-50 has 5 train/test splits of size 1.6K/400 respectively. 
DCASE2014~\citep{DCASE} is an acoustic scenes and event classification dataset which has 100 training and 100 testing sound clips spanning 10 different audio classes. 
We demonstrate competitive performance relative to the state-of-the- art, despite training on a much smaller and less audio-rich Kinetics-400 dataset. 
We extract 10 equally spaced 2-second sub-clips from each full audio sample of ESC- 50~\citep{esc50} and 60 1-second sub-clips from each full sample of DCASE2014~\citep{DCASE}. 
We save the activations that result from the audio encoder to quickly train the linear classifiers. 
We use activations after the last convolutional layer of the ResNet-9 and apply a max pooling with kernelsize (1,3) and stride of (1,2) without padding to the output. 
For both datasets, we then optimize a L2 regularized linear layer with batch size 512 using the Adam optimizer~\citep{kingma15adam} with learning rate $1$x$10^{-4}$, weight-decay set to $5$x$10^{-4}$ and the default parameters. 
The classification score for each audio sample is computed by averaging the sub-clip scores in the sample, and then predicting the class with the highest score. 
The mean top-1 accuracy is then taken across all audio clips and averaged across all official folds.

\subsection{Linear probing results}
\setlength{\tabcolsep}{3pt}
\begin{table}[tb]
\begin{center}
\begin{tabular}{l l l c c }
\toprule
	\textbf{Method}                              & \textbf{Architecture} & \textbf{Dataset} &
	\multicolumn{2}{c}{\textbf{Top-1 Acc\%}} \\
				                              &    &   & HMDB & UCF       \\
\midrule
	RotNet3D~\cite{jing2018self}              & S3D         & K600 & 24.8 & 47.7  \\
	CBT~\cite{sun2019contrastive}             & S3D+BERT    & K600 & 29.5 & 54.0  \\
	MemDPC~\cite{Han2020MemoryaugmentedDP}    & R-2D3D      & K400 & 30.5 & 54.1  \\
	AVSF~\cite{xiao2020audiovisual}           & AVSF        & K400 & 44.1 & 77.4  \\
	CoCLR~\cite{Han2020SelfsupervisedCF}      & S3D         & K400 & 46.1 & 74.5  \\
	\midrule
    \bf{Ours:  \methodname}                         & R(2+1)D-18  & K400  & \bf{48.2} & \bf{77.0}  \\
    \midrule
	MIL-NCE~\cite{miech2019endtoend}          & S3D         & HT    & 53.1 & 82.7  \\
	XDC~\cite{alwassel2019self}               & R(2+1)D-18  & IG65M & 56.0 & 85.3  \\
	MMV~\cite{alayrac2020selfsupervised}      & R(2+1)D-18  & AS    & 60.0 & 83.9  \\
	ELo~\cite{piergiovanni2020evolving}       & R(2+1)D-50  & Y8M   & 64.5 & --   \\
\bottomrule
\end{tabular}
\end{center}
\vspace{-1em}
\caption{\textbf{Comparison to state-of-the-art.} Transfer learning results on UCF-101 and HMDB-51 when video backbone is frozen. \label{tab:sota_linear}}
\end{table}
In~\cref{tab:sota_linear}, we compute the linear classification results of our model compared to other recent methods. 
We find that our best model has competitive 3-fold linear evaluation results of $48.2\%$ on HMDB-51 and $77.0\%$ on UCF-101.

\subsection{Supervised training on K-400}
Here we experiment with training supervisedly on Kinetics-400 and observing the effect of using feature cropping (with the configuration 2 medium and 2 small latent space crops). 
\begin{table}
\begin{center}

\begin{tabular}{cc}
\toprule
Fm-Crop & HMDB-51 Top-1 Acc. \\
\midrule
{\color{gray}\xmark} & 67.6 \\
\cmark               & 69.0 \\
\bottomrule
\end{tabular} 
\end{center} 
\vspace{0.5em}
\caption{\textbf{Supervised Training.} We train the R(2+1)D+Transformer architecture supervisedly on Kinetics-400 with and without feature crops enabled. \label{tab:ablation:supervised}}
\end{table}
The experimental results are given in \cref{tab:ablation:supervised}
We find that even though our method is designed for contrastive cross-modal pretraining, using feature crops can help in training in a supervised manner too.

\subsection{Audio-Visual Heatmap Visualizations}
\begin{figure}[h]
\center
\includegraphics[width=0.4\textwidth]{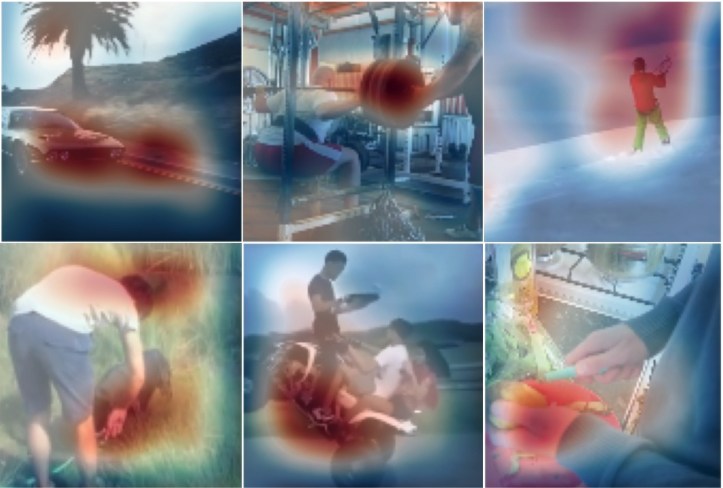}
\caption{\textbf{Heatmap visualizations.} 
Heatmaps are obtained by removing the spatial pooling layer and visualizing the strength of the dot-product of the audio feature vector with the video feature-map as in~\citep{arandjelovic2018objects}. 
Here, we show selected samples from Kinetics-400 training set of the resulting heatmaps along with the middle frame of the video.
}
\label{fig:heatmap}
\end{figure}

In \cref{fig:heatmap}, we show examples that our model truly learned some spatial correspondence between a region and audio. 
We have done this by visualizing the strength of the dot-product of the visual feature map (without pooling) with the audio feature vector.

\subsection{Preventing Shortcut Learning with Feature Crops.} 
Noise contrastive learning works better when you can reduce the mutual information between the input pairs~\citep{tian2020makes} as its harder for the network to cheat. 
This can be achieved by taking multiple spatial crops of images in the input space and independently applying different augmentations, such as color jittering and Gaussian blurring, to the cropped inputs.
However, as mentioned above, taking more than 2 crops in input space is both memory and computationally infeasible for multi-modal video data.
Crops in feature space, on the other hand, allows us to take multiple crops for noise contrastive learning. 
However, since CNNs have large receptive fields that easily cover the full frame, there may be shortcut learning with feature crops as information may leak between the crops from same feature map.
To alleviate this, we take feature crops from two originally augmented video clips, allowing us to make NCE comparisons \emph{across} modalities and individual augmentations (such as color jitter), leading to a beneficial reduction in mutual information.
Furthermore, while the theoretical receptive fields of units in later layers are indeed very large, units tend to be sensitive to an effective area which is significantly smaller than the theoretical receptive field~\citep{luo2017understanding, ZhouKLOT14}, further reducing the mutual information between inputs for noise contrastive learning.
\end{document}